\documentclass[10pt,twocolumn,letterpaper]{article}

\usepackage[pagenumbers]{iccv} 

\usepackage{amsmath,amsfonts,bm}

\def\eqref#1{equation~\ref{#1}}

\def\1{\bm{1}}

\def\vone{{\bm{1}}}

\def\vz{{\bm{z}}}

\def\mA{{\bm{A}}}
\def\mB{{\bm{B}}}
\def\mC{{\bm{C}}}

\def\mI{{\bm{I}}}

\def\mP{{\bm{P}}}
\def\mQ{{\bm{Q}}}
\def\mR{{\bm{R}}}
\def\mS{{\bm{S}}}

\def\mW{{\bm{W}}}
\def\mX{{\bm{X}}}

\DeclareMathAlphabet{\mathsfit}{\encodingdefault}{\sfdefault}{m}{sl}
\SetMathAlphabet{\mathsfit}{bold}{\encodingdefault}{\sfdefault}{bx}{n}

\def\gD{{\mathcal{D}}}

\definecolor{iccvblue}{rgb}{0.21,0.49,0.74}
\usepackage[pagebackref,breaklinks,colorlinks,allcolors=iccvblue]{hyperref}

\usepackage{xcolor,colortbl}

\usepackage{amsmath}
\usepackage{amssymb}
\usepackage{mathtools}

\usepackage{graphicx}
\usepackage{algorithmicx}
\usepackage{algorithm}
\usepackage{algpseudocode}
\usepackage{wrapfig}

\usepackage{multirow}
\usepackage{subcaption}
\usepackage{xcolor,colortbl}
\usepackage{siunitx}

\definecolor{lightred}{rgb}{1,0.8,0.8} 
\definecolor{lightgreen}{rgb}{0.8,1,0.8}
\definecolor{lightblue}{rgb}{0.88,0.96,1}
\definecolor{lightgrey}{rgb}{0.83, 0.83, 0.83}

\usepackage{enumitem} 

\usepackage{amsthm}
\usepackage{comment}

\newtheorem{proposition}{Proposition}

\theoremstyle{definition}

\newcommand{\lgr}{\cellcolor{lightblue}}

\usepackage{tikz}

\usepackage{pifont}
\newcommand{\cmark}{\color{iccvblue}\ding{51}}
\newcommand{\xmark}{\color{red}\ding{55}}

\usepackage{tocloft}

\def\paperID{1400} 
\def\confName{ICCV}
\def\confYear{2025}

\title{A Framework for Double-Blind Federated Adaptation of Foundation Models} 

\author{Nurbek Tastan\textsuperscript{1} \quad Karthik Nandakumar\textsuperscript{1,2} \\ 
\textsuperscript{1}Mohamed bin Zayed University of Artificial Intelligence (MBZUAI), UAE \\ 
\textsuperscript{2}Michigan State University (MSU), USA \\ 
{\tt\small nurbek.tastan@mbzuai.ac.ae, nandakum@msu.edu} 
}

\begin{document}
\maketitle
\begin{abstract}
    Foundation models (FMs) excel in zero-shot tasks but benefit from task-specific adaptation. However, privacy concerns prevent data sharing among multiple data owners, and proprietary restrictions prevent the learning service provider (LSP) from sharing the FM. In this work, we propose \textbf{BlindFed}, a framework enabling collaborative FM adaptation while protecting both parties: data owners do not access the FM or each other's data, and the LSP does not see sensitive task data. BlindFed relies on fully homomorphic encryption (FHE) and consists of three key innovations: (i) \textbf{FHE-friendly architectural modifications} via polynomial approximations and low-rank adapters, (ii) a \textbf{two-stage split learning} approach combining offline knowledge distillation and online encrypted inference for adapter training without backpropagation through the FM, and (iii) a \textbf{privacy-boosting scheme} using sample permutations and stochastic block sampling to mitigate model extraction attacks. Empirical results on four image classification datasets demonstrate the practical feasibility of the BlindFed framework, albeit at a high communication cost and large computational complexity for the LSP. The code can be found at \href{https://github.com/tnurbek/blindfed}{https://github.com/tnurbek/blindfed}.
\end{abstract}
    
\section{Introduction}
\label{sec: intro}

Foundation models (FMs) have transformed artificial intelligence, achieving state-of-the-art results across machine learning, computer vision, and natural language processing. Prominent examples include GPT \cite{radford2019language, brown2020language}, CLIP \cite{radford2021learning}, BERT \cite{devlin2018bert, he2020deberta, liu2019roberta}, Stable Diffusion \cite{rombach2022high}, Segment Anything \cite{kirillov2023segment}, and Vision Transformers \cite{dosovitskiy2020image, liu2021swin}. 
Though vision and multimodal FMs have demonstrated good zero-shot performance, there is often scope for performance improvement when faced with challenging out-of-domain tasks (e.g., medical images or satellite imagery). Hence, it becomes essential to adapt the FM for the downstream task. 

Adaptation of FMs for downstream tasks involves two main challenges: \textit{computational complexity} and \textit{data availability}. The simplest approach is transfer learning, where the FM serves as a frozen feature extractor, and only a classification head is trained -- linear probing is this head is a single linear layer. 
It is also possible to perform \textit{partial} (only a selected subset of parameters are adapted) or \textit{full finetuning} of the parameters of the FM based on the downstream data. 
Recent parameter-efficient fine-tuning (PEFT) methods fall into two categories: (i) \textit{prompt learning} \cite{jia2022visual}, which learns input or intermediate prompts without modifying FM parameters, and (ii) \textit{adapters} \cite{hu2021lora, dettmers2023qlora, mercea2024time}, which add trainable components to the FM. Adapters include sequential (e.g., low-rank adaptation a.k.a. LoRA \cite{hu2021lora}) and parallel (e.g., low-rank side adaptation a.k.a. LoSA \cite{mercea2024time}) variants. 
Except for transfer learning and parallel adapters, all the other adaptation techniques require partial or complete backpropagation through the FM, which is computationally expensive. 
\textit{Zeroth-order optimization} (ZOO) offers a backpropagation-free alternative for black-box FMs but incurs high cost due to numerous forward passes.

\begin{figure}
    \centering
    \includegraphics[width=\linewidth]{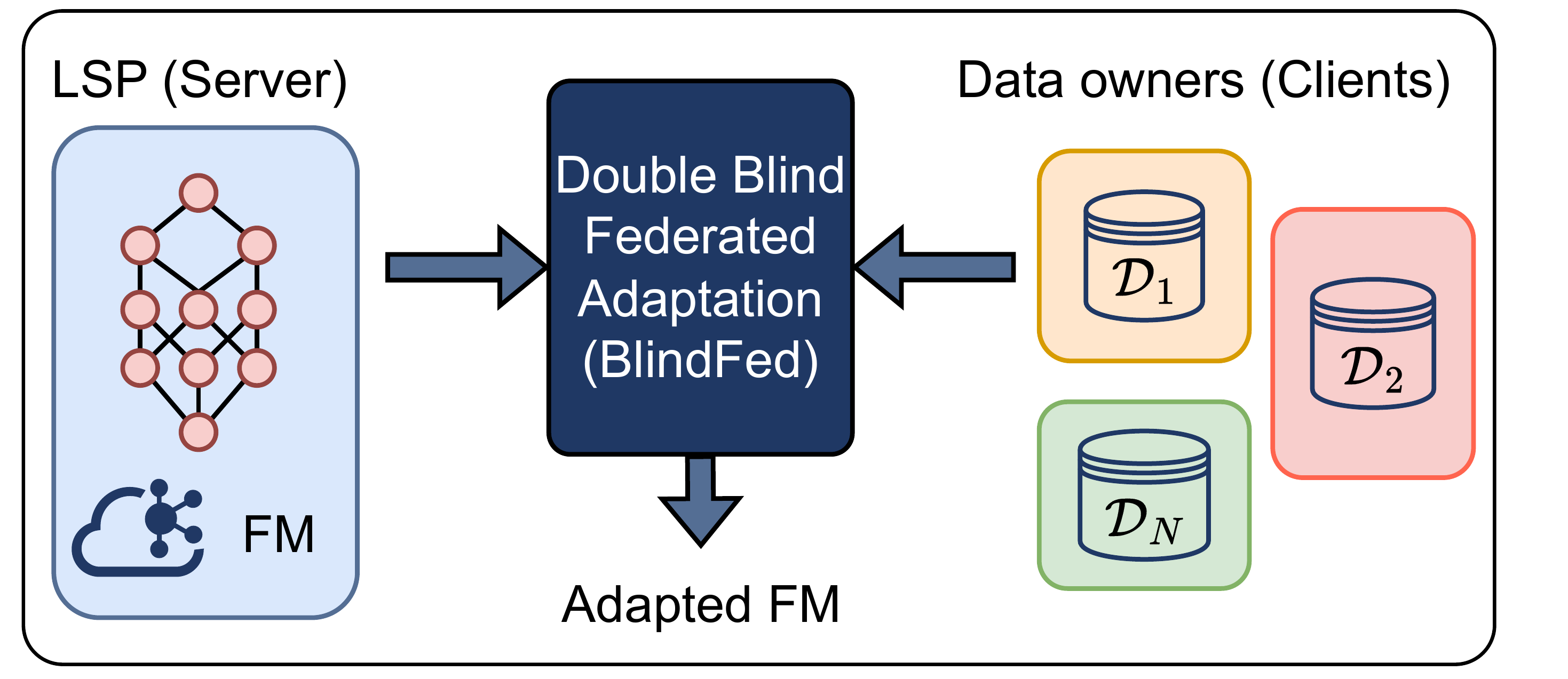} 
    \caption{Conceptual illustration of BlindFed framework for double-blind federated adaptation of a foundation model.}
    \label{fig: intro-figure}
\end{figure}

The other major challenge in FM adaptation is the unavailability of downstream training data to the learning service provider (LSP) who owns the FM. 
Moreover, this data may be distributed across multiple data owners (e.g., multiple hospitals or banks) and cannot be collated due to privacy concerns and regulations. 
Thus, FM adaptation requires collaboration between the LSP and data owners. Federated learning (FL) \cite{mcmahan2017communication} addresses this challenge by enabling collaborative training across entities while preserving data confidentiality. FL has been applied in many applications \cite{antunes2022federated, xu2021federated, long2020federated, liu2023efficient, sugianto2024collaborative, ghimire2022recent, allahham2024CLAP} and operates mainly in two settings: cross-silo (few data owners) and cross-device FL (large number of data owners) \cite{kairouz2021advances}. 

In this work, we focus on cross-silo federated adaptation of an FM (for an out-of-domain downstream image classification task) by an LSP (server) through collaboration with multiple data owners (clients) under two core constraints: (i) \textit{Model privacy} - the LSP wants to retain full ownership of the FM and does not want to share the FM with the data owners; and (ii) \textit{Data privacy} - clients do not want to reveal their data to the LSP or to each other. We jointly refer to these constraints as \textit{double-blind privacy} (see Figure \ref{fig: intro-figure}). We make the following four contributions:

\begin{itemize}
    \item We propose the BlindFed framework for double-blind federated adaptation of FMs based on well-known cryptographic tools such as fully homomorphic encryption (FHE) and secure multiparty computation (MPC). 
    \item We modify the given FM into an FHE-friendly architecture, leveraging existing ideas such as polynomial approximations and low-rank parallel adapters. 
    \item We propose a two-stage split learning approach, where the FHE-friendly FM blocks are first pre-trained via offline knowledge distillation, followed by online encrypted inference to train local parallel adapters, with MPC-based secure aggregation for the global adapter. 
    \item For stronger model privacy, we introduce a privacy boosting scheme based on sample-level permutation and stochastic block sampling.    
\end{itemize}

\section{Related Work}
\label{section: literature-review}

\begin{figure*}[t]
    \centering
    \includegraphics[width=\linewidth]{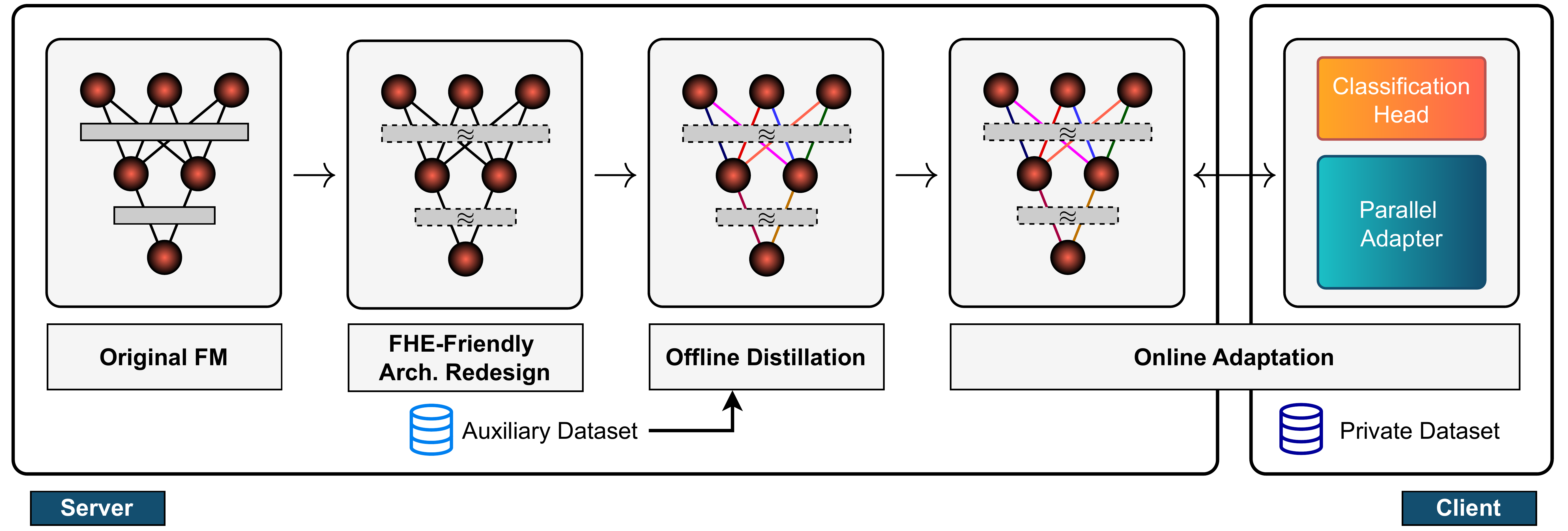}
    \caption{\textbf{Overview of the proposed BlindFed framework for double-blind federated adaptation.} The framework consists of three main components: (1) FHE-friendly architecture redesign, where the original foundation model (FM) is modified by approximating non-linear operations; (2) offline distillation, where the approximated blocks are fine-tuned via knowledge distillation using an auxiliary dataset; and (3) online adaptation, where clients interact the FHE-enabled FM under homomorphic encryption, performing local updates on the parallel adapter and classification head. } 
    \label{fig: main-scheme}
    \vskip -0.75em 
\end{figure*}

\noindent\textbf{Foundation models}:  
FMs have been highly successful in computer vision \cite{dosovitskiy2020image, radford2021learning, liu2021swin}, natural language processing \cite{radford2019language, liu2019roberta, yang2019xlnet, he2020deberta, clark2020electra}, and beyond \cite{sharir2021image, ranasinghe2022self}. In particular, the two-stage training strategy has shown to be effective, where FMs are first pre-trained on a large dataset for general understanding and then fine-tuned on a small downstream dataset to learn task-specific features. However, their vast scale introduces significant challenges, particularly in fine-tuning, that hinder their practical applicability. 

\noindent\textbf{Private inference}: The advent of machine learning as a service (MLaaS) has underscored the need for privacy-preserving techniques in ML, particularly in inference tasks. The concept of private inference (PI) has emerged as a pivotal solution to safeguard data and model privacy \cite{mohassel2017secureml, srinivasan2019delphi, shen2022abnn2, tan2021cryptgpu, zhang2023sal, li2022mpcformer, huang2022cheetah}. 
While vision transformers (ViTs) achieve strong performance, their high computational cost makes PI challenging, especially under cryptographic techniques such as FHE \cite{acar2018survey, gentry2009fully, cheon2017homomorphic} and MPC \cite{goldreich1998secure, knott2021crypten}. Most PI literature focuses on reducing computational and communication overheads while preserving accuracy. SAL-ViT \cite{zhang2023sal} improves ViT efficiency in PI, whereas Iron \cite{hao2022iron} optimizes matrix multiplication and key non-linear transformer operations (Softmax, GELU, LayerNorm). Another direction involves PI-friendly transformer designs, such as MPC-ViT \cite{zeng2023mpcvit}, which adapts ViTs for MPC with an accuracy-efficiency trade-off, and MPCFormer \cite{li2022mpcformer}, which combines MPC and knowledge distillation to reduce latency and maintain inference quality.

\noindent\textbf{Adaptation of foundation models}: 
The primary issue in adapting FMs is their massive size, making it impractical for individual users or clients with limited computational resources to fine-tune or even store them. Various PEFT techniques such as adapters \cite{houlsby2019parameter, lin2020exploring}, prompt learning \cite{li2021prefix}, low-rank adaptation (LoRA) \cite{hu2021lora}, and low-rank side adaptation (LoSA) \cite{mercea2024time} have been proposed. Numerous variants of LoRA, such as AdaLoRA~\cite{zhang2023adaptive}, Delta-LoRA~\cite{zi2023delta}, IncreLoRA~\cite{zhang2023increlora}, QLoRA~\cite{dettmers2023qlora}, LoRA-GA~\cite{wang2024loraga}, and LoFT~\cite{tastan2025loft} further optimize adaptation efficiency. 
These methods specifically target the transformer attention blocks, with LoRA modifying weight matrices to enable efficient fine-tuning with a lower computational load. However, LoRA still requires backpropagation through the backbone, increasing the total time taken to update the model. PEFT in federated settings has also been explored for LLMs~\cite{zhang2023fedpetuning, zhang2023towards}. 

\section{Problem Formulation}
\label{section: problem-formulation} 

Suppose that a foundation model (FM) $\mathcal{M}_{\psi}$ that is already pre-trained on a large-scale dataset is available with the learning service provider (LSP). The LSP aims to collaborate with the $K$ data owners to adapt the FM for a downstream image classification task. Each data owner $\mathcal{P}_k$ has access to a local training dataset $\mathcal{D}_k = \{\mathbf{x}_i^k,y_i^k\}_{i=1}^{N_k}$ corresponding to the downstream task. Here, $\mathbf{x}_i^k$ denotes the $i^{\text{th}}$ input image of $\mathcal{P}_k$, $y_i^k$ is the corresponding class label, $N_k$ is the number of training samples with $\mathcal{P}_k$, and $k \in [1,K]$. 

\noindent \textbf{Problem Statement}: Let $\widetilde{\mathcal{M}}_{\widetilde{\psi}}$ denote the FM adapted for the downstream task. The goal of the BlindFed framework is to collaboratively learn the parameters $\widetilde{\psi}$ under the following constraints: (i) \textit{Data Privacy}: the LSP does not learn anything about the local datasets $\{\mathcal{D}_k\}_{k=1}^{K}$ and the data owner $\mathcal{P}_k$ does not learn anything about other local datasets $\mathcal{D}_j$, where $j \neq k$; (ii) \textit{Model Privacy}: the data owners do not learn anything about the original FM $\mathcal{M}_{\psi}$.

\noindent \textbf{Assumptions}: To simplify the double-blind federated adaptation problem and make it practically feasible, we make the following assumptions: (i) \textit{Auxiliary dataset for preliminary adaptation}: We assume that the LSP has access to an independent auxiliary dataset $\mathcal{D}_{\text{aux}}$, which allows it to perform preliminary adaptation of the given FM into an image classifier. Note that this public dataset may not even correspond to the target image classification task. (ii) \textit{Modularity of FM}: We further assume that the FM has a modular architecture, which can be represented as a sequence of $L$ blocks, i.e., $\mathcal{M}_{\psi} = \mathcal{B}_{\psi_1} \circ \mathcal{B}_{\psi_2} \cdots \circ \mathcal{B}_{\psi_L}$. Specifically, a transformer architecture is considered in this work. (iii) \textit{Thin client}: The data owners do not have the resources to store the FM (or an encrypted version of it) and perform inference (or encrypted inference) using the FM. However, we assume that the data owners have sufficient computational resources to perform fully homomorphic encryption and decryption operations. (iv) \textit{Powerful server}: The LSP has enough computational resources to perform private inference on encrypted data transmitted by the data owners. Henceforth, we refer to the LSP and data owners as server and clients, respectively. (v) \textit{Semi-honest threat model}: Both the server and the clients are assumed to be semi-honest, i.e., they follow the adaptation protocol honestly, but may attempt to violate the privacy constraints. 

\noindent \textbf{Vanilla Federated Adaptation}: Let $\mathcal{L}_k(\widetilde{\psi}) = \frac{1}{N_k} \sum_{i=1}^{N_k} \mathbb{L}(\hat{y}_i^k,y_i^k)$ be the average loss at client $k$, where $\mathbb{L}$ denotes the per-sample loss and $\hat{y}_i^k = \widetilde{\mathcal{M}}_{\widetilde{\psi}}(\mathbf{x}_i^k)$ is the prediction output by the adapted model. Federated adaptation can be posed as a distributed optimization problem \cite{mcmahan2017communication}, where the goal is to learn the global model parameters $\widetilde{\psi}$ such that:
\begin{equation}
    \min_{\widetilde{\psi}} ~ \sum_{k=1}^K \alpha_k \mathcal{L}_k(\widetilde{\psi}), 
    \label{eq:fed}
\end{equation}
\noindent where $\alpha_k = \frac{N_k}{\sum_{j=1}^K N_j}$. In each round $t$ of FL adaptation, the server broadcasts the previous model parameters $\widetilde{\psi}^{(t-1)}$. Each client computes the local model parameters $\widetilde{\psi}_k^{(t)}$, and these local updates are aggregated by the server to obtain the current global model parameters $\widetilde{\psi}^{(t)}$. For example, simple FedAvg aggregation function can be represented as follows:
\begin{equation}
   \widetilde{\psi}^{(t)} = \sum_{k=1}^{K} \alpha_k \widetilde{\psi}_k^{(t)}.
   \label{eqn:FedAvg}
\end{equation}
\noindent Note that $t \in [1,T]$, where $T$ is the number of communication rounds and the model parameters $\widetilde{\psi}_k^{(0)}$ for the first round are typically initialized randomly by the server. 

\noindent \textbf{Challenges}: The above vanilla federated adaptation is not privacy-preserving because it requires computation of $\hat{y}_i^k = \widetilde{\mathcal{M}}_{\widetilde{\psi}}(\mathbf{x}_i^k)$, where the core FM $\mathcal{M}_{\psi}$ is available only at the server and the local training datasets are available only with the respective clients. Hence, it is essential to design a mechanism for computing $\widetilde{\mathcal{M}}_{\widetilde{\psi}}(\mathbf{x}_i^k)$ without violating the data and model privacy constraints. Moreover, the sharing of local updates $\widetilde{\psi}_k^{(t)}$ with the server could also potentially leak information about the local datasets \cite{zhu2019deep}. Hence, the aggregation step in Eq. \ref{eqn:FedAvg} must be performed securely without revealing the local updates to the server. 

\section{Proposed BlindFed Framework}
\begin{figure}[t]
    \centering
    \includegraphics[width=\linewidth]{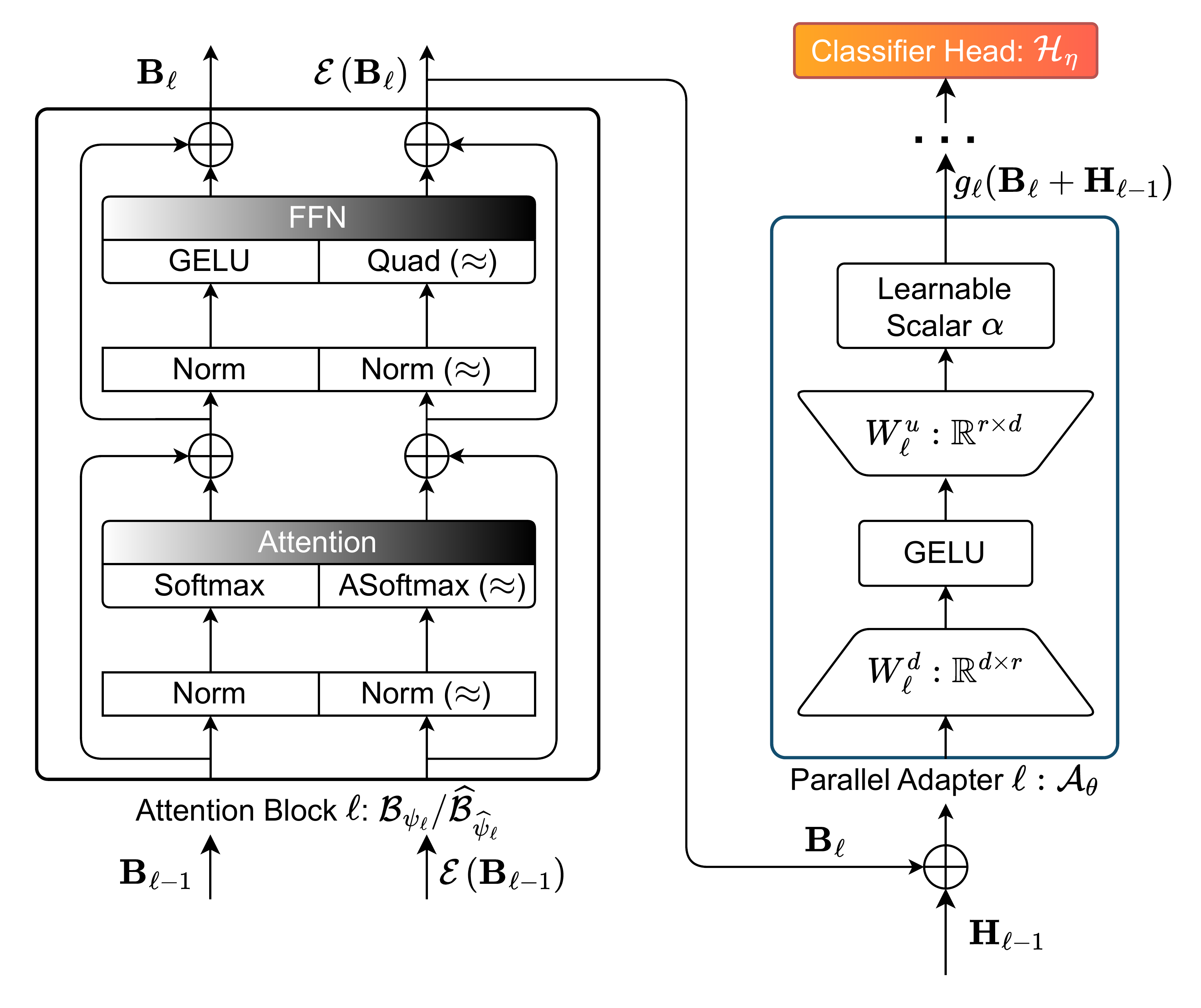}
    \caption{\textbf{FHE-friendly architecture redesign.} Each transformer block's non-linear operations -- GELU activations, Softmax attention, and the division step in LayerNorm -- are replaced with low-degree polynomial approximations (denoted ``Quad'' for GELU and ``ASoftmax'' for Softmax). A lightweight parallel adapter and classification head are then trained on the client side. } 
    \label{fig: main-full-diagram}
    \vskip -0.75em 
\end{figure}

The BlindFed framework for double-blind federated adaptation (Figure \ref{fig: main-scheme}) comprises two components: \textbf{\textit{(1) FHE-friendly Architecture Redesign}} -- The FM is first modified into an FHE-friendly model by approximating non-linear operations and adding a classification head as well as a parallel adapter; \textbf{\textit{(2) Two-stage Split Learning}} -- In the first \textit{offline} stage, the approximated individual blocks are fine-tuned through knowledge distillation from the original FM using an auxiliary dataset. In the second \textit{online} stage, clients encrypt their local data using FHE and interact with the server to perform encrypted inference. Based on intermediate outputs from the server, clients locally update the parallel adapter and classification head. The server then uses an MPC-based secure aggregation to combine these updates into global parameters, which are shared back with all clients. 
The overall training workflow of the proposed framework is summarized in Figure \ref{fig: main-full-diagram}. 
During inference, the encrypted inference step is repeated. Based on the intermediate outputs received from the server, the client utilizes the global parallel adapter and classifier to obtain the final class prediction. To enhance model privacy, the server can further incorporate sample-level permutations and stochastic block sampling. 

\subsection{FHE-friendly Architecture Redesign}

The first step in the proposed framework is to redesign the given FM into an FHE-friendly model by leveraging existing techniques. Assuming that the given FM follows a modular transformer encoder architecture with $L$ attention blocks (Figure~\ref{fig: losa-figure}), let $\mathbf{b}_{\ell-1}$ be the input to the $\ell^{\text{th}}$ attention block $\mathcal{B}_{\psi_\ell}$ and $\mathbf{b}_{\ell}$ be the corresponding output. We want to learn an FHE-friendly approximation of the block $\mathcal{B}_{\psi_\ell}$, denoted as $\widehat{\mathcal{B}}_{\widehat{\psi}_\ell}$, such that, for encrypted input $\mathcal{E}(\mathbf{b}_{\ell-1})$, the server computes encrypted output as $\mathcal{E}(\mathbf{b}_{\ell}) = \widehat{\mathcal{B}}_{\widehat{\psi}_\ell}(\mathcal{E}(\mathbf{b}_{\ell-1}))$, with the redesigned FM denoted as $\mathcal{\hat{M}}_{\hat{\psi}}$ consisting of a sequence of FHE-friendly blocks $\widehat{\mathcal{B}}_{\widehat{\psi}_\ell}$.

\noindent \textbf{Approximating Non-linear Functions}: 
Encrypted inference is limited to polynomial operations in most FHE schemes (e.g., CKKS \cite{cheon2017homomorphic}), requiring polynomial approximations for non-linear functions in transformer blocks: Softmax, GELU, and LayerNorm. In this work, Softmax is approximated using a Taylor series approximation of the exponential function $(e^x)$: 
\begin{equation}
    e^x = \sum_{i=0}^{\infty} \frac{x^i}{i!} \approx \sum_{i=0}^d \frac{x^i}{i!},  
    \label{eq: exp-approx}
\end{equation}
followed by normalization through division by the sum of the calculated exponential values. The error bound of this approximation is the remainder term, e.g. $\frac{e^{\xi}}{(d+1)!} x^{d+1}$, for some $\xi$ between $0$ and $x$. Furthermore, GELU activation is approximated via a simple quadratic function: 
\begin{equation}
    \text{GELU}(x) \approx \text{Quad}(x) = 0.125x^2 + 0.25x + 0.5. 
    \label{eq: gelu-approx} 
\end{equation}
The LayerNorm function and Softmax require a division, which is implemented via Goldschmidt's algorithm \cite{cheon2019numerical}: 
\begin{eqnarray}
    \frac{1}{x} = \frac{1}{1-(1-x)} &=& \prod_{i=0}^{\infty} \left(1+(1-x)^{2^i}\right) \nonumber \\ 
    &\approx& \prod_{i=0}^d \left( 1+(1-x)^{2^i} \right), \label{eq: division} 
\end{eqnarray}
where $x \in (0, 2)$. 

A task-specific classification head $\mathcal{H}_{\eta}$ and a parallel adapter $\mathcal{A}_{\theta}$ are appended to the approximated FM to enable adaptation. The choice of a parallel adapter is critical in the FHE-friendly redesign because sequential adapters like LoRA require backpropagation through the FM during adaptation, which is practically infeasible when the data remains encrypted. Thus, the redesigned FHE-friendly model can be considered a combination of the approximated FM, the parallel adapter, and the classifier, i.e., $\widetilde{\mathcal{M}}_{\widetilde{\psi}} = (\mathcal{\hat{M}}_{\hat{\psi}} || \mathcal{A}_{\theta}) \circ \mathcal{H}_{\eta}$, where $||$ indicates that these functions operate in parallel and $\circ$ is the composition operator. Though ideas such as the approximation of non-linear operations and a parallel adapter exist in the literature, we have carefully assembled these pieces to redesign the FM into an FHE-friendly model. 

\subsection{Two-stage Split Learning}

In the re-designed FHE-friendly FM, only the server stores the approximated FM; each client keeps a local parallel adapter and classifier. Training proceeds in two stages. 

\noindent \textbf{Stage 1: Offline Distillation.} Before any collaboration, the server trains the approximated FM (student) from the original FM (teacher) on the auxiliary dataset $\mathcal{D}_{aux}$. 
After replacing all non-linearities (Softmax, GELU, and Inverse) with their approximations, we distill four types of representations: (i) embeddings, (ii) attention matrices (pre-normalization), (iii) hidden states after each block, and (iv) final prediction layer \cite{jiao2019tinybert, hinton2015distilling, li2022mpcformer}. Following \cite{jiao2019tinybert}, the first half of epochs distills (i)-(iii); the second half distills (iv). Details of the distillation process appear in Appendix~\ref{section: how-distillation}.

\noindent \textbf{Stage 2: Online Adaptation.} This step is performed via an interactive protocol between the clients and the server, which can be further divided into three phases: (i) encrypted inference, (ii) local learning, and (iii) secure aggregation.

\subsubsection{Encrypted Inference}
\label{subsec:EncryptedInference}

FMs exceed the multiplicative depth supported by current FHE schemes; evaluating the whole network homomorphically would incur large approximation errors or require frequent (and impractical) bootstrapping, especially under the thin client assumption. Hence, we propose performing encrypted inference only over a single transformer block at a time. After each block, the client decrypts and re-encrypts the intermediate representation $\mathcal{E}(\mathcal{F}(\mathcal{E}(\mathbf{b}_{\ell})))$, and returns it back to the server. Here, $\mathcal{F}$ is the decryption operation. 

The overall encrypted inference protocol can be summarized as follows. At the beginning of the collaboration, each client $\mathcal{P}_k$ encrypts (using its public key) its local inputs and labels $\{\mathcal{E}(\mathbf{x}_i),\mathcal{E}(y_i)\}_{i=1}^{N_k}$ and sends them to the server. The server applies the embedding function on the encrypted data to obtain the input to the first attention layer $\{\mathcal{E}(\mathbf{b}_0^i)\}_{i=1}^{N_k}$. 
Subsequently, for each FL round, the server randomly selects a batch of $n$ samples from this set, say $\mathcal{E}(\mathbf{B}_{0}) = [\mathcal{E}(\mathbf{b}_{0}^1),\mathcal{E}(\mathbf{b}_{0}^2),\cdots,\mathcal{E}(\mathbf{b}_{0}^n)]$, and sequentially performs encrypted inference on each FHE-friendly block $\widehat{\mathcal{B}}_{\widehat{\psi}_{\ell}}$. After block $\ell$, the client decrypts these representations (using its private key), re-encrypts them again (using its public key), and returns them to the next block. 

When the client receives the output of the final transformer attention block $\mathcal{E}(\mathbf{B}_{L})$, the decrypted representations are passed through the classification head and the final predictions are again encrypted to get $\mathcal{E}(\mathbf{\hat{Y}}) = [\mathcal{E}(\hat{y}^1), \mathcal{E}(\hat{y}^2), \cdots, \mathcal{E}(\hat{y}^n)]$. These encrypted predictions are sent back to the server for per-sample loss computation in the encrypted domain. The server computes the batch loss and sends this encrypted average loss to the client. The client decrypts this loss and uses it for local learning. Throughout, all the operations on the server are carried out in the encrypted domain. Since the server does not have access to the client's private key, the server learns no information about the client's local data. On the other hand, the client receives a batch of intermediate representations (in plaintext) after each attention block. 

\subsubsection{Local Learning}
\label{subsec:LocalLearning}
\begin{figure}
    \centering
    \includegraphics[width=\linewidth]{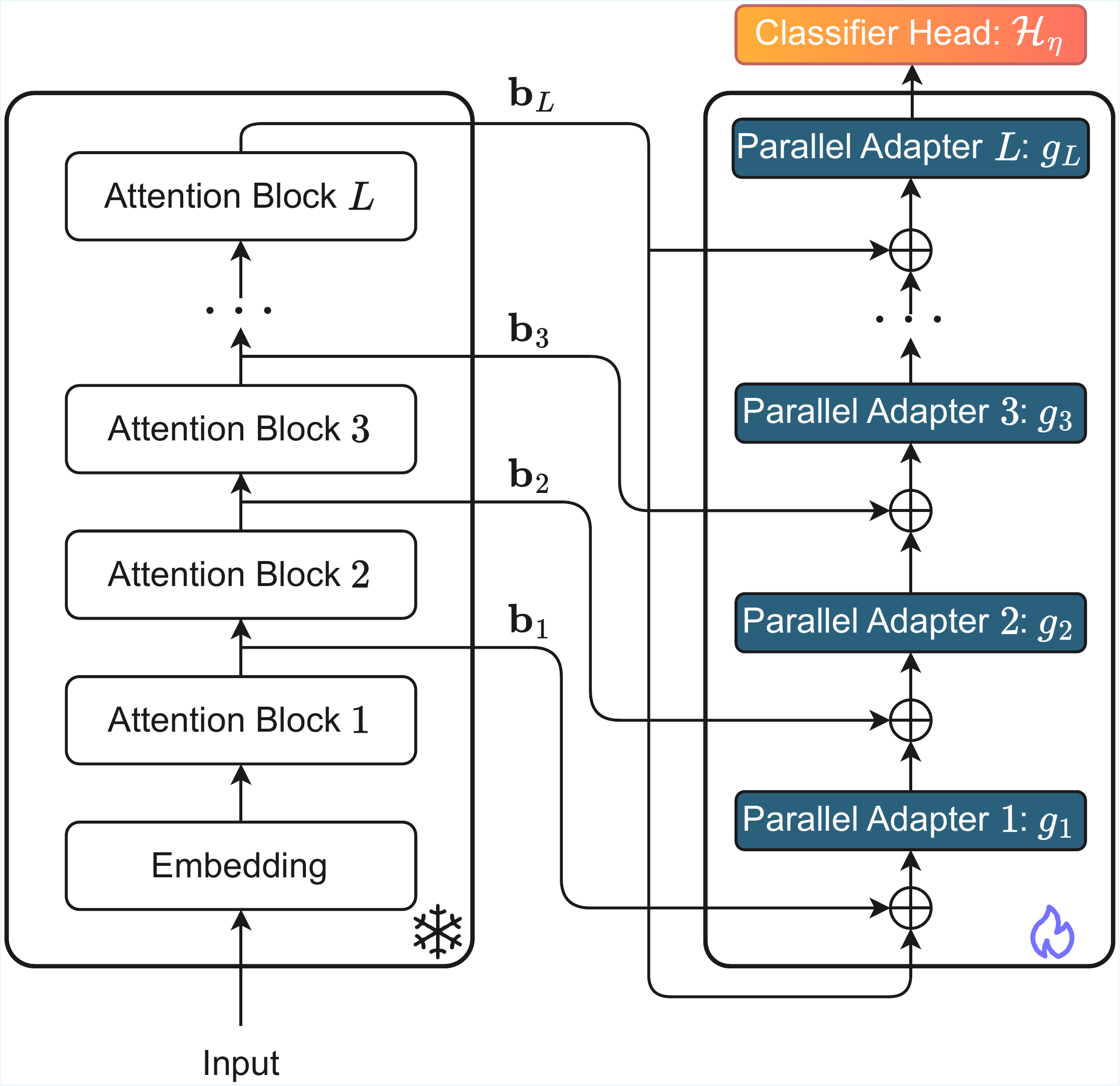}
    \caption{Illustration of the parallel adapter design.} 
    \label{fig: losa-figure}
    \vskip -0.75em 
\end{figure}

Though various model adaptation strategies are available, the proposed framework requires an adaptation method that does not require backpropagation of gradients through the FM. This leaves us with only two possible choices: \textit{transfer learning} (where only the classification head is learned) and \textit{parallel adapter} (where both the classification head and a side adapter are learned). In this work, we adopt the low-rank parallel adapter method proposed in \cite{mercea2024time} (see Figure~\ref{fig: losa-figure}). This method requires access to intermediate representations after every transformer attention block, which is readily available (in plaintext) through the encrypted inference protocol described in Section \ref{subsec:EncryptedInference}. 

The output of the low-rank parallel adapter corresponding to the attention block $\ell$ can be expressed as:
\begin{equation}
    \mathbf{h}_{\ell} = g_{\ell}(\mathbf{b}_{\ell} + \mathbf{h}_{\ell-1}) + \mathbf{h}_{\ell-1},
    \label{eqn:overallAdapterOutput}
\end{equation}
\noindent where $\mathbf{h}_{0} = \mathbf{b}_L$. The adapter function $g_{\ell}$ is given by:
\begin{equation}
    g_{\ell}(\mathbf{z}) = \alpha \mathbf{W}_{\ell}^u \text{GELU}(\mathbf{W}_{\ell}^d\mathbf{z}),
    \label{eqn:adapterfn}
\end{equation}
\noindent where $\mathbf{W}_{\ell}^d$ and $\mathbf{W}_{\ell}^u$ are the down- and up-projection matrices and $\alpha_{\ell}$ is the scaling factor at block $\ell$. Finally, the client locally updates the parallel adapter and classification head in the plaintext domain based on the average loss received from the server, employing the same procedure as in \cite{poirot2019split}. 

\subsubsection{Secure Aggregation} 
To ensure the secure aggregation of parameter updates from clients, we leverage secure multi-party computation (MPC)~\cite{bonawitz2017practical}. This approach enables the aggregation server to compute the average of client updates without gaining access to the individual updates themselves. In BlindFed, the local adapters and classifiers are securely aggregated via FedAvg to obtain the global adapter and classification head. 

\subsection{Model Privacy Boosting}

The downsides of performing encrypted inference over one attention block at a time are two-fold. Firstly, it increases the communication cost because encrypted intermediate outputs are exchanged between the client and the server after every block in every FL round. Since communication efficiency is not one of our core constraints, we consider the increased communication cost as a limitation, not a red flag. However, since the intermediate representations $\mathbf{b}_{\ell}$ after every attention block are accessible to the client in plaintext form, a malicious client could use $(\mathbf{b}_{\ell-1},\mathbf{b}_{\ell})$ pairs for multiple training samples to mount a model extraction attack~\cite{liang2024model} and learn the parameters of each transformer block. This clearly violates the model privacy constraint. To circumvent this problem and preserve model privacy, we introduce two changes to the online adaptation stage, namely, \textit{sample-level permutation} and \textit{stochastic block sampling}. 

\subsubsection{Sample-level Permutation}
Each communication round processes a batch of samples. Let $\mathcal{E}(\mathbf{B}_{\ell}) = [\mathcal{E}(\mathbf{b}_{\ell}^1),\mathcal{E}(\mathbf{b}_{\ell}^2),\cdots,\mathcal{E}(\mathbf{b}_{\ell}^n)]$ be a batch of encrypted intermediate representations corresponding to a client, where $n$ is the batch size. Before sending these representations to the client, the server applies a $n \times n$ permutation matrix $\mathbf{\Pi}_{\ell}$ and sends only the permuted batch $\mathcal{E}(\mathbf{B}_{\ell})\cdot \mathbf{\Pi}_{\ell} = [\mathcal{E}(\mathbf{b}_{\ell}^{\pi(1)}),\mathcal{E}(\mathbf{b}_{\ell}^{\pi(2)}),\cdots,\mathcal{E}(\mathbf{b}_{\ell}^{\pi(n)})]$ to the client. Here, $[\pi(1),\pi(2),\cdots,\pi(n)]$ represents a random permutation of $[1,2,\cdots,n]$. This permutation matrix $\mathbf{\Pi}_{\ell}$ can be randomly selected for each block $\ell$ in each communication round. Thus, the client never sees corresponding pairs $(\mathbf{b}_{\ell-1}^i,\mathbf{b}_{\ell}^i)$ for any training sample $i$ in the batch, ensuring some protection against model extraction attacks. 

Because the adapter in Eq.~\ref{eqn:adapterfn} is applied per sample, the permutation of samples within a batch does not affect this computation. However, the adapter output in Eq. \ref{eqn:overallAdapterOutput} depends on values from two consecutive blocks, which have undergone different permutations. Hence, it is necessary to ensure consistent permutation of the inputs. When operating on a batch of samples, Eq. \ref{eqn:overallAdapterOutput} can be reformulated as: 
\begin{equation}
    \mathbf{H}_{\ell} = g_{\ell}(\mathbf{B}_{\ell} + \mathbf{H}_{\ell-1}) + \mathbf{H}_{\ell-1},
    \label{eqn:overallAdapterOutputBatch}
\end{equation}
\noindent where $\mathbf{H}_{0} = \mathbf{B}_L$. Note that the client receives only a permutation of intermediate representations, i.e., $(\mathbf{B}_{\ell} \cdot \Pi_{\ell})$ and not the original $\mathbf{B}_{\ell}$, $\forall~ \ell \in [1,L]$. Hence, to facilitate the computations associated with the parallel adapter, the server also sends $(\Pi_{\ell-1}^{-1} \cdot \Pi_{\ell})$ for all $\ell \in [2,L]$ as well as $(\Pi_{L}^{-1} \cdot \Pi_{1})$ to the client. When the client receives $(\mathbf{B}_{L} \cdot \Pi_{L})$, it can compute $\mathbf{H}^{'}_{0} = (\mathbf{B}_L \cdot \Pi_{L}) \cdot (\Pi_{L}^{-1} \cdot \Pi_{1}) = (\mathbf{B}_L \cdot \Pi_{1}) = (\mathbf{H}_{0} \cdot \Pi_{1})$. This can be directly used in Eq. \ref{eqn:overallAdapterOutputBatch} along with $(\mathbf{B}_{1} \cdot \Pi_{1})$ to compute $\mathbf{H}^{'}_{1} = (\mathbf{H}_{1} \cdot \Pi_{2})$. Following the same logic, it is possible to compute $\mathbf{H}^{'}_{\ell} = (\mathbf{H}_{\ell} \cdot \Pi_{\ell+1}), ~ \forall~ \ell \in [1,L]$. When the server receives the final encrypted predictions from the client, it can permute the encrypted labels of the batch using $\Pi_{L+1}$ before computing the per-sample losses and aggregating them. It must be emphasized that revealing $(\Pi_{\ell-1}^{-1} \cdot \Pi_{\ell})$ for all $\ell \in [2,L]$ as well as $(\Pi_{L}^{-1} \cdot \Pi_{1})$ to the client does not leak any information about $\Pi_{\ell}$ as shown in Proposition \ref{lemma: 1}. 

\begin{proposition}
    Let $\mA, \mB,$ and $\mC$ be $n \times n$ permutation matrices. Given only $\mA^{-1}\mB$, $\mB^{-1}\mC$, and $\mC^{-1}\mA$, it is computationally infeasible to uniquely recover the individual matrices $\mA$, $\mB$, and $\mC$ without additional information. 
    \label{lemma: 1}
\end{proposition}

\begin{proof}[Proof of Proposition \ref{lemma: 1}]
    See Appendix \ref{appendix: proof-of-proposition}.

\end{proof}

\vspace{-1.5em}
\subsubsection{Stochastic Block Sampling (SBS)} 
\label{section: sbs}

Sample-level permutation ensures that samples in each batch are randomly permuted based on $\mathbf{\Pi}_{\ell}$. Because the client never sees $\mathbf{\Pi}_{\ell}$, it is not straightforward to mount a model extraction attack when the batch size $n$ is sufficiently large. However, intermediate representations of the same sample by two successive transformer blocks are likely to have higher similarity than representations from different samples (Appendix \ref{appendix: feature-sim-attack}). This similarity can be exploited to recover individual permutation matrices or, at the very least, reduce the brute-force search space. To mitigate this risk, we introduce stochastic block sampling (SBS): at each server-side forward pass, we return only a subset of block outputs and set the rest to zero, so the full sequence of representations is never revealed. 

\begin{wrapfigure}{r}{0.46\linewidth}
    \centering
    \vspace{-1em}
    \includegraphics[width=\linewidth]{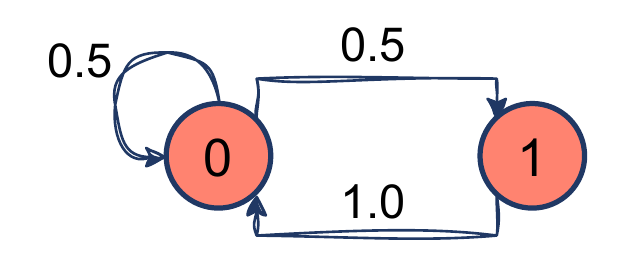}
    \vspace{-2em}
    \caption{Stochastic block sampling strategy.} 
    \label{fig: stoch-sampling}
    \vspace{-1em}
\end{wrapfigure}

A key consideration in this strategy is avoiding the sampling of consecutive (neighboring) blocks, as this could still enable similarity-based attacks. As shown in Figure~\ref{fig: stoch-sampling} (and Appendix \ref{appendix: feature-sim-attack}), feature similarity is negligible when blocks are separated by at least one layer. We therefore use a structured sampling process: 
(i) if block $\ell$ is sampled (state $1$), the next block $\ell+1$ is not sampled (probability $1$); (ii) if block $\ell$ is not sampled (state $0$), the next block $\ell+1$ is sampled with a probability of $0.5$. Thus, the proposed model privacy boosting techniques ensure that the encrypted inference protocol is double-blind, even though the intermediate representations are exposed in plaintext form.

\begin{table*}[th]
    \centering
    \resizebox{\linewidth}{!}{ 
        \begin{tabular}{llcccccc} 
            \toprule 
             \multirow{2.5}{*}{\textbf{Datasets}} & \multirow{2.5}{*}{\textbf{Methods}} & \multirow{2.5}{*}{\parbox{2cm}{\centering \textbf{Is double- \\ blind? }}} & \textbf{Centralized} & & \multicolumn{3}{c}{\textbf{Federated} $(K=5)$} \\ \cmidrule{4-4} \cmidrule{6-8} 
             & & & \textbf{Pooled} & & \textbf{Dirichlet} $(\alpha=100)$ & \textbf{Dirichlet} $(\alpha=1)$ & \textbf{Dirichlet} $(\alpha=0.01)$ \\ \midrule 
             \multirow{6.5}{*}{\textbf{CIFAR-10}} 
             & Full fine-tuning                   & \xmark & $0.9635$ & & $0.9759$ & $0.9725$ & $0.8857$ \\ 
             & LoRA                               & \xmark & $0.9592$ & & $0.9736$ & $0.9718$ & $0.8979$ \\ 
             & Adapter tuning                     & \xmark & $0.8992$ & & $0.8681$ & $0.8539$ & $0.6754$ \\ \cmidrule{2-8} 
             & Linear probing                     & \cmark & $0.9226$ & & $0.9203$ & $0.9191$ & $0.7447$ \\ 
             & \lgr \textbf{BlindFed} & \lgr\cmark & \lgr $0.9428$ & \lgr& \lgr$0.9471$ & \lgr$0.9413$ & \lgr$\mathbf{0.8540}$ \\ 
             & \lgr \textbf{BlindFed + SBS} & \lgr\cmark & \lgr $\mathbf{0.9443}$ & \lgr & \lgr$\mathbf{0.9486}$ & \lgr$\mathbf{0.9427}$ & \lgr $0.8489$ \\ \midrule 
             \multirow{6.5}{*}{\textbf{CIFAR-100}} 
             & Full fine-tuning                   & \xmark & $0.8361$ & & $0.8684$ & $0.8611$ & $0.7882$ \\ 
             & LoRA                               & \xmark & $0.8349$ & & $0.8593$ & $0.8568$ & $0.7647$ \\ 
             & Adapter tuning                     & \xmark & $0.6594$ & & $0.6495$ & $0.6396$ & $0.4489$ \\ \cmidrule{2-8} 
             & Linear probing                     & \cmark & $0.7476$ & & $0.7486$ & $0.7414$ & $0.5317$ \\ 
             & \lgr \textbf{BlindFed} & \lgr\cmark & \lgr$\mathbf{0.7930}$ &\lgr & \lgr$\mathbf{0.7929}$ & \lgr$\mathbf{0.7808}$ & \lgr$\mathbf{0.6620}$ \\ 
             & \lgr \textbf{BlindFed + SBS} & \lgr\cmark & \lgr$0.7869$ & \lgr & \lgr$0.7861$ & \lgr$0.7789$ & \lgr$0.6584$ \\ \midrule 
             \multirow{6.5}{*}{\textbf{SVHN}} 
             & Full fine-tuning                   & \xmark & $0.9680$ & & $0.9763$ & $0.9692$ & $0.7601$ \\ 
             & LoRA                               & \xmark & $0.9659$ & & $0.9656$ & $0.9709$ & $0.7545$ \\
             & Adapter tuning                     & \xmark & $0.5201$ & & $0.5251$ & $0.4785$ & $0.3325$ \\ \cmidrule{2-8} 
             & Linear probing                     & \cmark & $0.5938$ & & $0.5879$ & $0.5732$ & $0.3385$ \\ 
             & \lgr \textbf{BlindFed} & \lgr\cmark & \lgr$0.9232$ &\lgr & \lgr$\mathbf{0.9329}$ & \lgr$0.9249$ & \lgr$0.7431$ \\ 
             & \lgr \textbf{BlindFed + SBS} & \lgr\cmark & \lgr$\mathbf{0.9257}$ & \lgr & \lgr$0.9298$ & \lgr$\mathbf{0.9256}$ & \lgr$\mathbf{0.7434}$ \\ \bottomrule 
        \end{tabular}
    }
    \caption{Comparison of accuracy achieved by our proposed method against baseline approaches on three datasets (CIFAR-10, CIFAR-100, and SVHN) in both centralized and federated learning scenarios. Federated experiments involve five clients $(K=5)$ with data partitioned using a Dirichlet distribution at varying levels of heterogeneity $(\alpha=100, 1, 0.01)$. The best-performing results among double-blind algorithms are highlighted in bold.} 
    \label{tab: main-comparison-acc} 
    \vskip -0.75em 
\end{table*}

\label{section: proposed-approach} 

\section{Experiments and Results} 
\label{section: experiments}

\subsection{Datasets} 

To validate and implement our approach, we utilize a well-known vision transformer (ViT) pretrained on ImageNet-1K (ViT-Base) as the FM. 
The public auxiliary datasets for distilling the FHE-friendly FM are: \textbf{Tiny-Imagenet} \cite{le2015tiny} is a subset of the larger ImageNet dataset, containing 200 different classes, each with 500 images, 100 validation/test images, totaling 120K images. \textbf{Fed-ISIC2019} \cite{terrail2022flamby} is a multi-class dataset of dermoscopy images containing $23247$ images across $8$ melanoma classes, with high label imbalance. This dataset provides an opportunity to test our framework within the healthcare domain. For the distillation phase of our experiments, we exclusively use data from client 1. 

\noindent The private downstream datasets are: \textbf{CIFAR-10 and CIFAR-100} \cite{krizhevsky2009learning} datasets are standard benchmarks for image classification, containing $60000$ color images across $10$ and $100$ classes, respectively. \textbf{SVHN} \cite{netzer2011reading} dataset is a benchmark for digit recognition, consisting of over $73257$ images of house numbers for training and $26032$ for testing. \textbf{Fed-ISIC2019.} The remaining data points for centers 2-6 are used in the fine-tuning experiments, aligning well with the federated setup, as the dataset is tailored for federated learning. For the centralized setup, all data points are pooled into a single client.

\subsection{Experimental Setup}
We employ the Vision Transformer (ViT) \cite{dosovitskiy2020image}, pre-trained on the ImageNet-1k dataset \cite{russakovsky2015imagenet} (ViT-Base), with a backbone dimension of $384 \times 384$. For the first phase of our framework, obtaining the FHE-friendly FM, we use Adam optimizer with a learning rate of $10^{-4}$ for distilling the transformer blocks for 15 epochs and $10^{-5}$ for distilling the prediction layer for the remaining 15 epochs, totaling 30 epochs. We set the batch size to $16$ due to the substantial memory demands. We use MSE loss for the first phase of the distillation and the combination of cross-entropy loss and Kullback-Leibler (KL) divergence losses for the second phase. We set the polynomial order of exponential approximation to $6$ and the order of the inverse to $7$. 
For the second phase of our framework, federated adaptation, we use SGD optimizer with a learning rate of $0.001$ for linear probing and our proposed method and $5\cdot10^{-5}$ for full fine-tuning experiments. We set the total number of communication rounds to $T=50$, and we use a learning rate scheduler with a decay factor of $0.1$ at rounds $[25, 40]$. We set the batch size to $16$ unless otherwise specified. We use cross-entropy loss to evaluate the effectiveness of the global model and report balanced accuracy. We use Dirichlet distribution-based splitting for all our experiments except the Fed-ISIC2019 dataset, which is naturally partitioned. 
All our experiments are conducted on NVIDIA A100-SXM4-40GB GPUs on an internal cluster server, with each run utilizing a single GPU.

\begin{table}[th]
    \vskip 0.5em 
    \centering
    \resizebox{\linewidth}{!}{ 
    \begin{tabular}{llccc}
        \toprule 
        \textbf{Public dataset} & \textbf{Methods} & \textbf{Is DB?} & \textbf{Centralized} & \textbf{Federated} \\ \midrule 
        \multirow{6.5}{*}{\parbox{2.25cm}{\centering \textbf{Fed-ISIC2019 \\ (center=0) \\ (InD)}}} 
        & Full fine-tuning                              & \xmark & $0.7811$ & $0.6752$ \\ 
        & LoRA                                          & \xmark & $0.7347$ & $0.6844$ \\
        & Adapter tuning                                & \xmark & $0.6601$ & $0.5762$ \\ \cmidrule{2-5}
        & Linear probing                                & \cmark & $0.6599$ & $0.5856$ \\ 
        & \lgr \textbf{BlindFed}                            & \lgr\cmark & \lgr $0.7090$ & \lgr $0.6679$ \\ 
        & \lgr \textbf{BlindFed + SBS}                      & \lgr\cmark & \lgr $\mathbf{0.7169}$ & \lgr $\mathbf{0.6831}$\\ \midrule 
        \multirow{6.5}{*}{\parbox{2.25cm}{\centering \textbf{Tiny-Imagenet \\ (OOD)}}} 
        & Full fine-tuning                              & \xmark & $0.7817$ & $0.6985$ \\ 
        & LoRA                                          & \xmark & $0.7330$ & $0.6880$ \\ 
        & Adapter tuning                                & \xmark & $0.6702$ & $0.6074$ \\ \cmidrule{2-5}
        & Linear probing                                & \cmark & $0.6372$ & $0.5789$ \\ 
        & \lgr \textbf{BlindFed}                            & \lgr\cmark & \lgr $0.7051$ & \lgr $0.6481$ \\ 
        & \lgr \textbf{BlindFed + SBS}                      & \lgr\cmark & \lgr $\mathbf{0.7127}$ & \lgr $\mathbf{0.6581}$ \\ \bottomrule 
    \end{tabular}
    }
    \caption{Performance comparison of our method with baseline approaches on the Fed-ISIC2019 dataset with five clients ($K=5$), using two auxiliary datasets: Fed-ISIC2019 (center=0) as an in-distribution (InD) dataset and Tiny-ImageNet as an out-of-distribution (OOD) dataset. DB refers to Double-Blind.} 
    \label{tab: fedisic}
    \vskip -1em 
\end{table}

\subsection{Results} 

\textbf{Main results: } In Table \ref{tab: main-comparison-acc}, the performance of our proposed method and baseline methods across three datasets (CIFAR-10, CIFAR-100, and SVHN) is evaluated under both centralized and federated learning settings. The federated learning experiment uses a Dirichlet partitioning strategy with varying levels of data heterogeneity, controlled by the Dirichlet concentration parameter $\alpha$ (ranging from $100$ to $0.01$). The results demonstrate that full fine-tuning achieves the highest accuracy across all datasets and settings, particularly excelling in more homogeneous federated scenarios, but it is computationally expensive and not double-blind. Linear probing maintains reasonable performance in homogeneous settings but fails drastically on SVHN and under extreme heterogeneity, confirming its limitations in adaptation. Our approach delivers robust and competitive performance, closely aligning with LoRA in accuracy while maintaining significantly lower computational demands (see Appendix \ref{appendix: efficiency-results}). Among the model privacy boosting techniques, sample-level permutation does not have any impact on the accuracy of the adapted model, but SBS may affect local learning because of missing intermediate representations. However, in practice, BlindFed+SBS demonstrates comparable performance to BlindFed without SBS, suggesting that SBS has minimal impact on adapted model performance while boosting model privacy. In some cases, these missing values add robustness to the learning process, leading to marginally better generalization performance.

\noindent\textbf{Fed-ISIC2019 results: } Table \ref{tab: fedisic} compares the performance of our method and baselines on the Fed-ISIC2019 dataset with five centers. The auxiliary datasets used at the LSP include (1) Fed-ISIC2019 with only the first center (treated as an in-distribution dataset) and (2) Tiny-ImageNet (treated as an out-of-distribution dataset). The results demonstrate that knowledge transfer from the OOD dataset is effective for all the methods, highlighting that the auxiliary dataset used for offline distillation can be any available dataset.

\begin{table}[t] 
    \centering
    \resizebox{\linewidth}{!}{ 
    \begin{tabular}{lcccc}
        \toprule 
         \textbf{Methods} & \textbf{Is DB?} & $K=10$ & $K=20$ & $K=50$ \\ \midrule 
         Full fine-tuning & \xmark & $0.9739$ & $0.9513$ & $N/A$ \\ 
         LoRA & \xmark & $0.9661$ & $0.9584$ & $0.9482$ \\ 
         Adapter tuning & \xmark & $0.8696$ & $0.8494$ & $0.8165$ \\ \midrule
         Linear probing   & \cmark & $0.9167$ & $0.9142$ & $0.9007$ \\  
         \rowcolor{lightblue} \textbf{BlindFed} & \cmark & $\mathbf{0.9446}$ & $\mathbf{0.9422}$ & $0.9287$ \\ 
         \rowcolor{lightblue} \textbf{BlindFed + SBS} & \cmark & $0.9425$ & $0.9411$ & $\mathbf{0.9388}$ \\ 
         \bottomrule 
    \end{tabular}
    }
    \caption{Scalability analysis of the proposed method to baseline approaches on the CIFAR-10 dataset, with varying number of clients $K \in \{10, 20, 50\}$ under a Dirichlet concentration parameter of $1.0$ for data partitioning. $N/A$ - one GPU is insufficient to run the experiment. DB refers to Double-Blind.} 
    \label{tab: number-of-participants}
    \vskip -0.75em 
\end{table}

\noindent\textbf{Scalability Results:} Table \ref{tab: number-of-participants} illustrates the scalability of our method and the baseline approaches on the CIFAR-10 dataset with an increasing number of clients $(K=\{10, 20, 50\})$ using Dirichlet partitioning $(\alpha=1.0)$ and a fixed batch size of 8. Full fine-tuning achieves the highest accuracy for $K = 10$ and $K = 20$ but becomes infeasible for $K = 50$ due to GPU limitations (we use only one GPU for each of our experiments). Linear probing demonstrates stable performance, but our method outperforms linear probing in all the settings, balancing compute efficiency, scalability, and accuracy, and demonstrating its suitability for federated setups with a large number of clients. 

\noindent\textbf{Communication Overhead:} In standard FL (FedAvg~\cite{mcmahan2017communication}), the communication cost depends on the foundation model (FM) size. In our work, ViT-Base is used as a FM consisting of ${\approx}86$M parameters requiring ${\approx}344$MB of bandwidth. In practice, LSPs often deploy larger FMs, e.g., ViT-Huge or even bigger models (${\approx}22$B parameters), which require ${\approx}88$GB of bandwidth. In contrast, BlindFed requires the transmission of an encrypted intermediate representation (IR) for each transformer block. In our work, an IR is a $577 \times 768$ tensor, which requires $\mathbf{6.21 MB}$ in plaintext and $C{=}\mathbf{17.33 MB}$ after encryption (${\approx}\mathbf{2.8}\times$ expansion) (see Appendix~\ref{appendix: homomorphic}). Ignoring the tiny adapter update ($0.25$M params. ${\approx}1$MB), the total communication cost of BlindFed is $(N_k * L * C)$, transmitted in batches, where $N_k$ is the local training dataset size and $L$ is the no. of transformer blocks. For federated adaptation tasks, $N_k$ is expected to be small, and not all IRs need to be transmitted for SBS. Hence, the communication overhead of BlindFed will be significantly higher compared to FedAvg for smaller FMs but becomes comparable when adapting large FMs with limited local data (which is often the case in practice). 

\noindent Other ablation studies and computational complexity analyses are reported in Appendix~\ref{appendix: add-experimental-results}.

\section{Conclusions}

This paper offers a promising framework for adapting foundation models for critical downstream applications without compromising on the data confidentiality and model privacy. However, the BlindFed framework is just a first step and needs to be improved in many dimensions before it is ready for practical adoption. Firstly, the high communication cost of BlindFed and high computational complexity at the server need to be mitigated. Secondly, more rigorous model privacy guarantees would be required before LSPs can expose valuable proprietary models to collaborators. Finally, the robustness of the framework in the presence of malicious collaborators should be carefully analyzed.       

\newpage 

\section*{Acknowledgments} 

This material is partly based on work supported by the Office of Naval Research N00014-24-1-2168.

{
    \small
    \bibliographystyle{ieeenat_fullname}
    \bibliography{main}
}

\clearpage

\onecolumn 
\tableofcontents

\maketitlesupplementary

\appendix 

\section{Proof of Proposition \ref{lemma: 1}}
\label{appendix: proof-of-proposition}

\begin{proof}[Proof of Proposition \ref{lemma: 1}]
    Let $\mP = \mA^{-1}\mB$, $\mQ = \mB^{-1}\mC$, and $\mR=\mC^{-1}\mA$, where $\mP$, $\mQ$, and $\mR$ are known products derived from matrices $\mA$, $\mB$, and $\mC$. Each of $\mA, \mB,$ and $\mC$ is a permutation matrix, defined by: 
    \begin{equation}
        \begin{aligned}
            \mA, \mB, \mC &\in \mathbb{P}_n = \big\{ \mX \in \{0,1\}^{n \times n}: \\ 
            & \mX^T\mX=\mI, \mX \vone = \vone \big\}, 
        \end{aligned}
    \end{equation}
    where $\mathbb{P}_n$ denotes the set of $n \times n$ permutation matrices, $\mX^T$ is the transpose of $\mX$, $\vone$ is a vector of ones, and $\mX^T=\mX^{-1}$. 

    \paragraph{Non-uniqueness of solutions:} To recover $\mA$, $\mB$, and $\mC$ uniquely, we would need the products $\mP$, $\mQ$, and $\mR$ to uniquely determine a single set of matrices $(\mA, \mB, \mC)$. However, for any valid solution $(\mA, \mB, \mC)$ that satisfies $\mP = \mA^{-1}\mB$, $\mQ = \mB^{-1}\mC$, and $\mR=\mC^{-1}\mA$, we can construct alternative solutions by applying a left multiplication with any permutation matrix $\mS \in \mathbb{P}_n$. Define alternative matrices $\mA^{\prime} = \mS\mA$, $\mB^{\prime} = \mS\mB$, and $\mC^{\prime} = \mS\mC$, then: 
    \begin{equation}
        \begin{aligned}
            \mA^{\prime -1} \mB^{\prime} &= (\mS\mA)^{-1} (\mS\mB) \\ 
            &= \mA^{-1}\mS^{-1} \mS\mB \\ 
            &= \mA^{-1}\mB = \mP, 
        \end{aligned}
    \end{equation}
    and similarly, $\mB^{\prime -1}\mC^{\prime} = \mQ$ and $\mC^{\prime -1}\mA^{\prime} = \mR$. 
    Thus, the products $\mP$, $\mQ$, and $\mR$ are also satisfied by the set $(\mA^{\prime}, \mB^{\prime}, \mC^{\prime})$. Since there are $n!$ possible choices for $\mS$, there are $n!$ distinct sets of matrices $(\mA^{\prime}, \mB^{\prime}, \mC^{\prime})$ that satisfy the products $\mP$, $\mQ$, and $\mR$. 

    \paragraph{Brute-force as an optimal attack strategy:} Given the non-uniqueness property above, an attacker aiming to recover $\mA$, $\mB$, and $\mC$ must resort to brute force, testing all possible configurations of $(\mA, \mB, \mC)$ that satisfy the products $(\mP, \mQ, \mR)$. The total number of combinations of $(\mA, \mB, \mC)$ the attacker would need to test is $n!$ and for $n=16$, we need $16! \approx 2 \cdot 10^{13}$ combinations. 
    
\end{proof}

\section{Feature Similarity-based Attack} 
\label{appendix: feature-sim-attack}

A potential attack strategy involves matching the $(\mathbf{b}_{\ell}, \mathbf{b}_{\ell+1})$ pairs, with $\ell \in [L-1]$, based on their similarity (e.g., L2 distance). After clients decrypt the received permuted representations $\mathbf{b}_{\ell}$, they could attempt to identify approximate nearest representations between the neighboring block outputs, thereby aiding in recovering individual permutation matrices $\Pi_{\ell}$. 

\begin{figure}
    \centering
    \includegraphics[width=\linewidth]{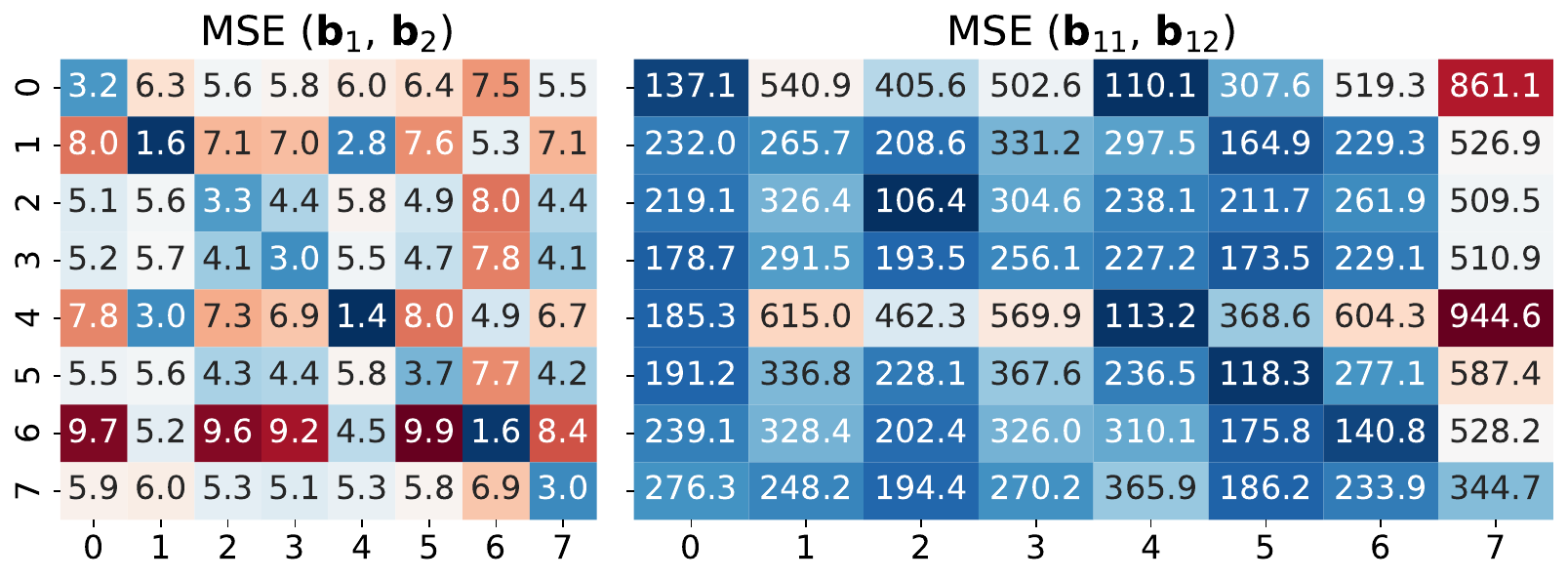}
    \caption{Heatmap visualization of the similarity (L2 distance) between corresponding block outputs or hidden representations: (i) between block outputs $\mathbf{b}_1$ and $\mathbf{b}_2$, and (ii) between block outputs $\mathbf{b}_{11}$ and $\mathbf{b}_{12}$. The X- and Y-axes represent the samples, with a batch size of $8$. An ideal outcome for the attacker would be a perfect minimal diagonal, indicating strong alignment between the representations. While this is not fully observed in the plot, such outcome would reduce the number of brute-force trials required, as outlined in Proposition \ref{lemma: 1}.} 
    \label{fig: mse-consecutive-blocks} 
\end{figure}

\begin{figure}
    \centering
    \includegraphics[width=\linewidth]{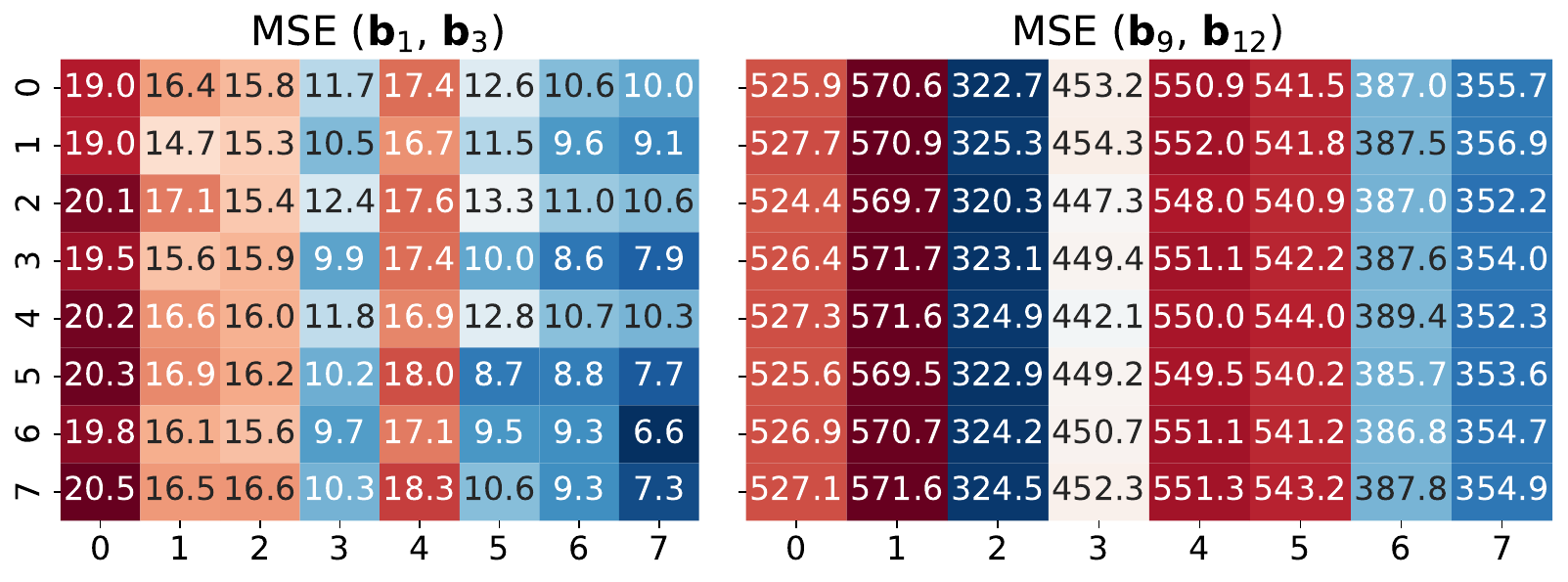}
    \caption{Heatmap visualization of the similarity (L2 distance) between non-consecutive block outputs or hidden representations: (i) between block outputs $\mathbf{b}_1$ and $\mathbf{b}_3$, and (ii) between block outputs $\mathbf{b}_{9}$ and $\mathbf{b}_{12}$. The X- and Y-axes represent the samples, with a batch size of $8$. }
    \label{fig: mse-random-blocks}
\end{figure}

To demonstrate this, we provide Figure \ref{fig: mse-consecutive-blocks}, which illustrates that neighboring blocks exhibit similarity, making it feasible to infer permutation mappings. However, if clients receive randomly sampled blocks instead of sequential ones, the same attack strategy fails (see Figure \ref{fig: mse-random-blocks}). To mitigate this risk, we introduce stochastic sampling of hidden representations/attention blocks in every forward pass. This ensures that the permutation matrices are never fully exposed, preventing adversaries from reconstructing them (see Proposition \ref{lemma: 1}). 

\begin{figure*}
    \centering
    \includegraphics[width=\linewidth]{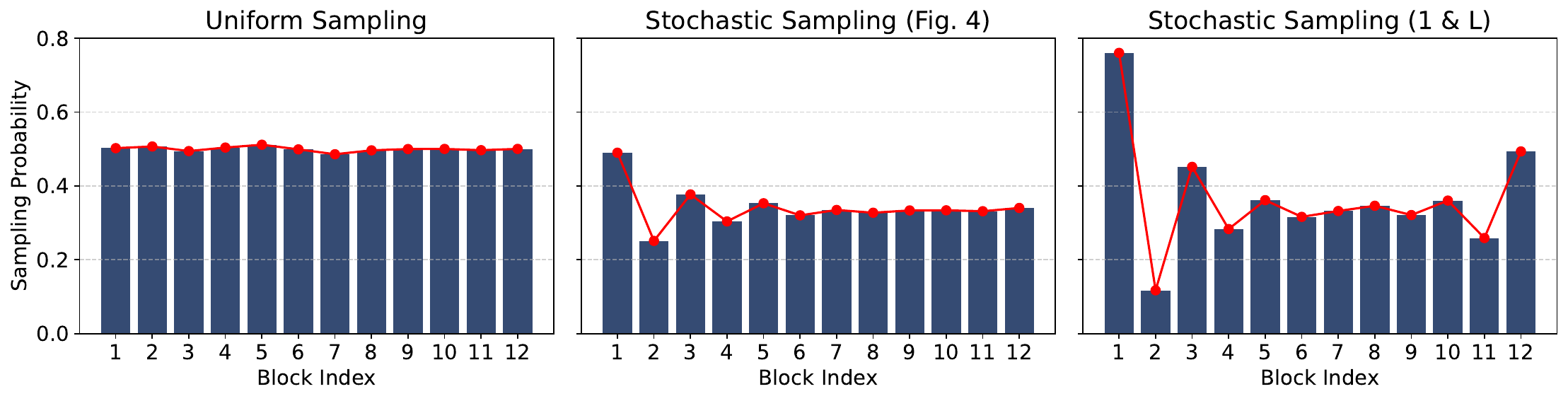}
    \caption{Visualization of the sampling probabilities for $L=12$ transformer blocks under three different scenarios: (i) uniform random sampling with probability $0.5$, (ii) our proposed stochastic sampling strategy, and (iii) stochastic sampling constrained to include either the first or last block.} 
    \label{fig: sampling-probs}
\end{figure*}

Additionally, we illustrate the sampling probabilities of transformer blocks under different strategies in Figure \ref{fig: sampling-probs}. Specifically, we compare uniform sampling (included for illustrative purposes), our proposed stochastic sampling strategy depicted in Figure \ref{fig: stoch-sampling}, and a constrained version of this stochastic strategy where either the first or last transformer block must always be sampled. The latter strategy ensures meaningful local updates by guaranteeing non-zero input entries for clients. Despite this constraint, the overall expected number of sampled blocks remains unchanged and aligns with the formulation in Eq. \ref{eq: total-expected-number-of-blocks}. Also, the learnable scaling parameter in each adapter layer (see Figure \ref{fig: main-full-diagram}), denoted by $\alpha$, addresses potential scaling issues that may arise due to zeroed-out block outputs/hidden representations during stochastic block sampling.

\subsection{HE-Compatible Strategy}
To maintain privacy while adhering to the computational constraints of homomorphic encryption (HE), we employ the following strategy: 
\begin{itemize}
    \vspace{1em}
    \item[--] Non-selected blocks are set to zeros, ensuring they are not used during the local learning. 
    \item[--] Due to limited multiplicative depth in HE, we still transmit the non-selected blocks but introduce random noise before sending them to clients. 
    \item[--] Clients decrypt, re-encrypt, and return the updates. The server then subtracts the added noise before proceeding to the next block, ensuring that adversaries cannot infer the original representations. 
    \vspace{1em}
\end{itemize}

This defense mechanism effectively counters feature similarity-based attacks while maintaining the efficiency of our privacy-preserving adaptation framework. A detailed explanation of how sampling is done is provided in Section~\ref{section: sbs}.

\subsection{Complexity Analysis of the Sampling Strategy} 
Consider a sequence consisting of $L$ consecutive blocks, each assigned either a value of $1$ (sampled) or $0$ (not sampled), according to the probabilistic rules depicted in Figure~\ref{fig: stoch-sampling}: 
\begin{itemize}
    \vspace{1em}
    \item The first block is sampled (assigned the value $1$) with probability $0.5$ and not sampled (assigned $0$) with probability $0.5$. 
    \item For each subsequent block $\ell \geq 2$: 
    \begin{itemize}
        \item If block $\ell-1$ was sampled, then block $\ell$ is deterministically set to $0$. 
        \item If block $\ell-1$ was not sampled, then block $\ell$ is sampled/not sampled with probability $0.5$. 
    \end{itemize}
\end{itemize}

Let $p_{\ell}$ denote the probability of block $\ell$ being sampled. Formally, this probability can be expressed recursively as: 
\begin{equation}
    p_{\ell} = 0.5 \times (1 - p_{\ell-1}), \quad \text{ with } p_1 = 0.5. 
\end{equation}

\paragraph{Expected Number of Sampled Blocks.}
Solving this recurrence relation reveals an equilibrium behavior for large $L$. Specifically, as $L$ gets larger, $p_{\ell}$ converges to a stationary probability $p$, satisfying: 
\begin{equation}
    p = 0.5 (1-p). 
\end{equation}
Solving this yields the stationary probability $p=\frac{1}{3}$. Therefore, each block has approximately a $\frac{1}{3}$ probability of being sampled. For $L$ blocks, the expected number of sampled blocks $S$ can be expressed as: 
\begin{equation}
    \mathbb{E}[S] = \sum_{\ell=1}^{L} p_{\ell} \approx \frac{L}{3}.
\end{equation}

When the sampling is repeated for $T$ times, the total expected number of sampled blocks is given by: 
\begin{equation}
    \mathbb{E}[S_T] \approx T \times \frac{L}{3}. 
    \label{eq: total-expected-number-of-blocks}
\end{equation}

\section{How Does the Distillation Work?} 
\label{section: how-distillation}
We adopt the transformer distillation approach \cite{jiao2019tinybert}, which consists of two key phases: (i) transformer-layer distillation and (ii) prediction-layer distillation. For generality, let the original transformer model be referred to as the teacher model $(\mathcal{T})$ and the approximation-enabled transformer model as the student model $(\mathcal{S})$. We assume that both models share the same architecture but differ only in their non-linear components (Softmax, GELU, LayerNorm). Further, we outline the primary sub-layers of the transformer blocks: 
\begin{itemize}
    \item \textbf{Attention.} The attention function is formulated as follows: 
    \begin{equation}
        \mathbf{A} = \frac{\mathbf{Q}\mathbf{K}^T}{\sqrt{d_k}},  
        \label{eq: attention}
    \end{equation}
    followed by a softmax operation. We are specifically interested in an un-normalized attention matrix $\mathbf{A}$. 
    \item \textbf{Feed-forward network} is formulated as follows: 
    \begin{equation}
        \mathbf{H}(x) = \text{GELU}(x\mW_1 + b_1)\mW_2 + b_2. 
        \label{eq: hidd}
    \end{equation}
\end{itemize}

\begin{figure}[t]
    \centering
    \includegraphics[width=\linewidth]{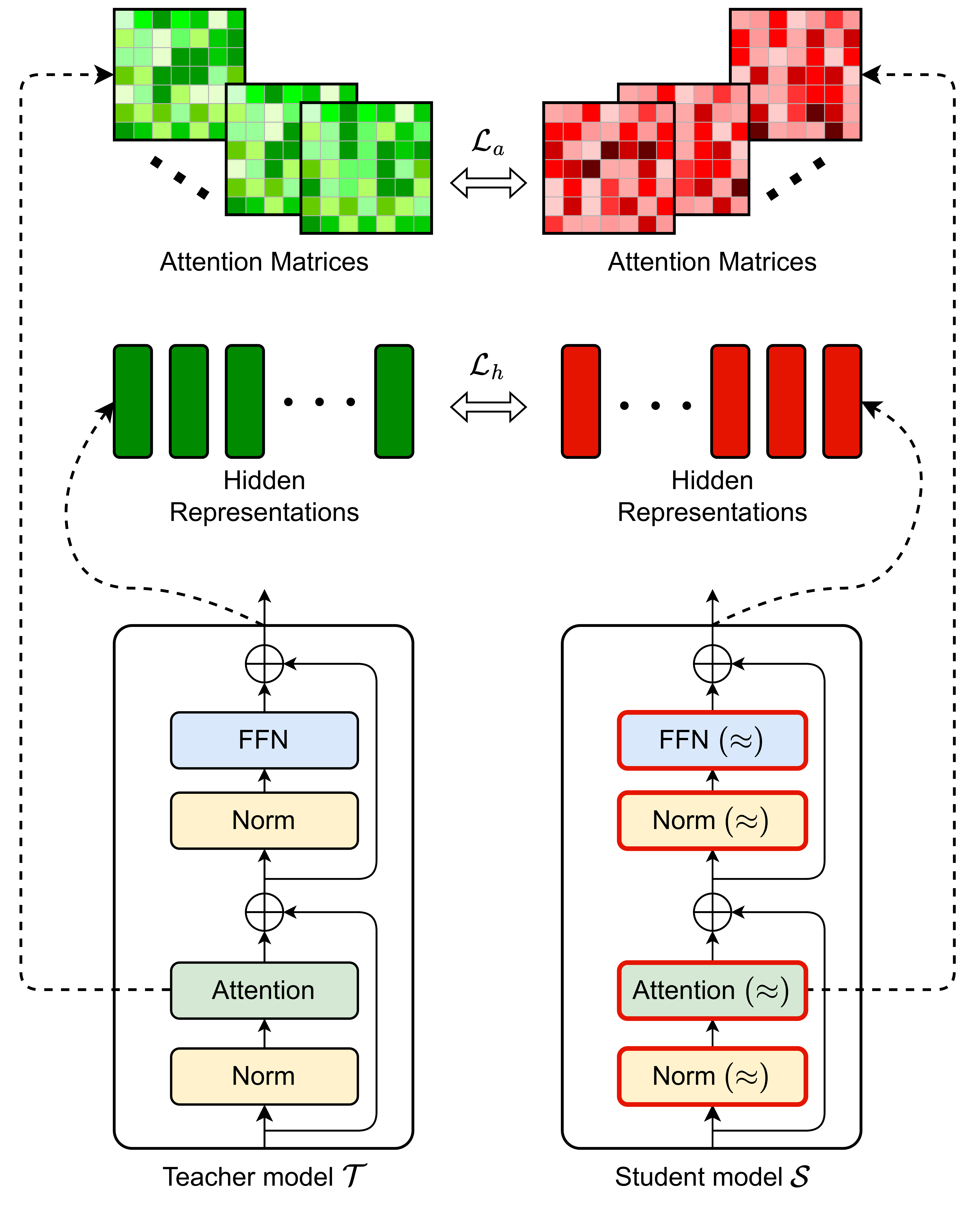}
    \caption{Schematic illustration of the layer-wise knowledge distillation (Equations \ref{eq: attn} and \ref{eq: hidd2}).}
    \label{fig: distillation}
\end{figure}

\begin{table*}[t]
    \centering
    \resizebox{0.925\linewidth}{!}{ 
        \begin{tabular}{llcrrrrr}
            \toprule 
             \textbf{Datasets} & \textbf{Methods} & \textbf{Is double-blind?} & \multicolumn{2}{c}{\textbf{No. of params (M)}} & \multicolumn{2}{c}{\textbf{Latency (ms)}} & \textbf{Memory (GB)} \\ \midrule 

             \multirow{5.5}{*}{\textbf{CIFAR-10 / SVHN}} 
             & Full fine-tuning                   & \xmark & $82.1096$ & $(11247.89\ \times)$ & $47.4$ & $(3.14\ \times)$ & $18.13$ \\ 
             & LoRA                               & \xmark & $0.2886$  & $(39.54 \times)$     & $38.5$ & $(2.55 \times)$  & $15.73$ \\ 
             & Adapter tuning                     & \xmark & $3.3933$  & $(464.68 \times)$    & $37.9$ & $(2.51 \times)$ & $14.72$ \\ \cmidrule{2-8} 
             & Linear probing                     & \cmark & $0.0073$  & $(1.00\ \times)$     & $15.1$ & $(1.00\ \times)$ & $9.39$ \\ 
             & \lgr \textbf{BlindFed} & \lgr\cmark & \lgr $0.2536$  & \lgr $(34.74\ \times)$    & \lgr $15.7$ & \lgr $(1.04\ \times)$ & \lgr $9.08$ \\ \midrule 

             \multirow{5.5}{*}{\textbf{CIFAR-100}} 
             & Full fine-tuning                   & \xmark & $82.1756$ & $(1121.09\ \times)$  & $47.4$ & $(3.03\ \times)$ & $18.13$ \\ 
             & LoRA                               & \xmark & $0.3546$  & $(4.84 \times)$  & $38.4$ & $(2.45 \times)$ & $16.42$ \\ 
             & Adapter tuning                     & \xmark & $3.4593$  & $(47.19 \times)$ & $37.8$ & $(2.41 \times)$ & $15.41$ \\ \cmidrule{2-8} 
             & Linear probing                     & \cmark & $0.0733$  & $(1.00\ \times)$     & $15.7$ & $(1.00\ \times)$ & $9.40$ \\ 
             & \lgr \textbf{BlindFed} & \lgr\cmark & \lgr $0.3196$  & \lgr $(4.36\ \times)$ & \lgr $16.3$ & \lgr $(1.04\ \times)$ & \lgr $9.08$ \\ \midrule 

             \multirow{5.5}{*}{\textbf{Fed-ISIC2019}} 
             & Full fine-tuning                   & \xmark & $82.1082$ & $(13916.64\ \times)$ & $51.2$ & $(2.54\ \times)$ & $17.32$ \\ 
             & LoRA                               & \xmark & $0.2871$  & $(48.66 \times)$  & $38.5$ & $(1.91 \times)$ & $15.73$ \\ 
             & Adapter tuning                     & \xmark & $3.3919$  & $(574.89 \times)$ & $37.9$ & $(1.88 \times)$ & $14.72$ \\ \cmidrule{2-8} 
             & Linear probing                     & \cmark & $0.0059$  & $(1.00\ \times)$     & $20.2$ & $(1.00\ \times)$ & $8.71$ \\ 
             & \lgr \textbf{BlindFed} & \lgr\cmark & \lgr $0.2522$ & \lgr $(42.75\ \times)$ & \lgr $21.1$ & \lgr $(1.05\ \times)$ & \lgr $8.39$ \\ \bottomrule 
        \end{tabular}
    }
    \caption{Comparison of the efficiency of our method with baseline approaches in terms of the number of parameters, latency and the memory requirement across four datasets (CIFAR-10/SVHN, CIFAR-100, and Fed-ISIC2019). Latency is measured per data point and includes both forward and backward passes. The last column (Memory (GB)) indicates the GPU memory required to conduct the experiment.} 
    \label{tab: main-comparison-params} 
\end{table*}

In the first stage of distillation, we perform an attention-based distillation (Eq. \ref{eq: attention}) and hidden representations based distillation (Eq. \ref{eq: hidd}). More precisely, 
\begin{equation}
    \mathcal{L}_{a} = \frac{1}{h} \sum_{i=1}^h \left\|\mathbf{A}_i^{\mathcal{S}} - \mathbf{A}_i^{\mathcal{T}}\right\|^2, 
    \label{eq: attn}
\end{equation}
where $h$ is the number of attention heads. And the distillation of the hidden representation is formulated as follows: 
\begin{equation}
    \mathcal{L}_{h} = \|\mathbf{H}^{\mathcal{S}} - \mathbf{H}^{\mathcal{T}}\|^2, 
    \label{eq: hidd2}
\end{equation}
where the matrices $\mathbf{H}^{\mathcal{S}}$ and $\mathbf{H}^{\mathcal{T}}$ are hidden representations of the respective models. For a detailed look, please refer to Figure~\ref{fig: distillation}, which illustrates how layer-wise distillation works. Then, the total loss of the first stage is simply defined as: 
\begin{equation}
    \mathcal{L} = \mathcal{L}_{a} + \mathcal{L}_{h}. 
\end{equation}

\begin{figure*}
    \centering
    \includegraphics[width=\linewidth]{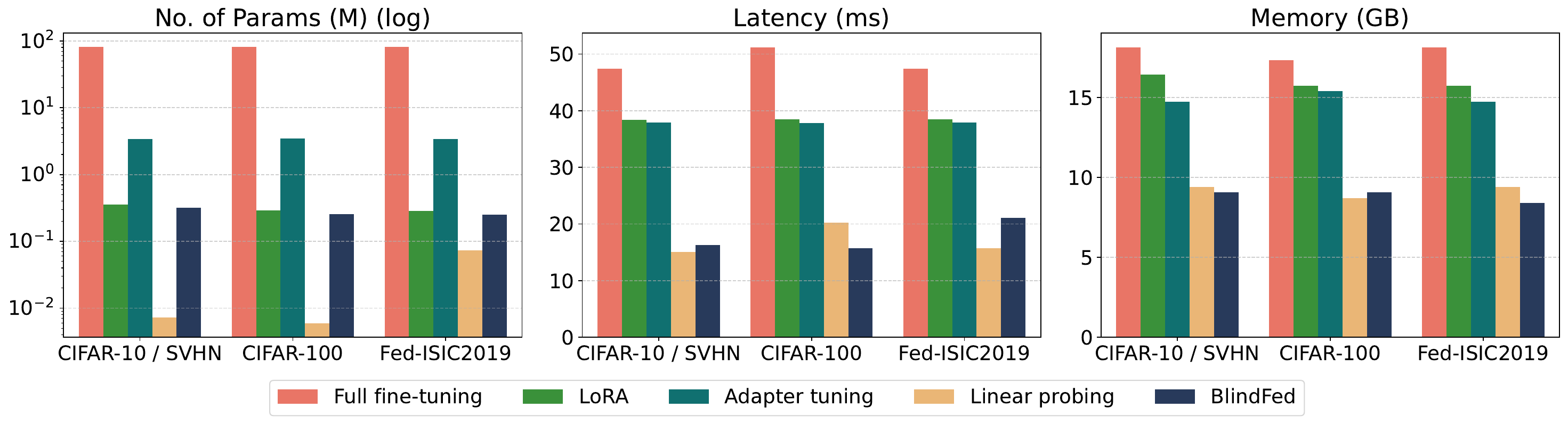}
    \caption{Comparison of the efficiency of different adaptation methods in terms of the number of parameters, latency, and memory usage across three datasets: CIFAR-10/SVHN, CIFAR-100, and Fed-ISIC2019. Each plot presents one of the three key metrics, with bars grouped by dataset and color-coded by method. Latency is measured per data point and includes both forward and backward passes. The memory column indicates the GPU memory required for training. Our proposed BlindFed method achieves a favorable balance between efficiency and performance. The results are derived from Table \ref{tab: main-comparison-params}. } 
    \label{fig: efficiency-barplots} 
\end{figure*}

Further, we perform a prediction-layer distillation following the knowledge distillation approach of \cite{hinton2015distilling} by matching logits and using the following objective function: 
\begin{equation}
    \mathcal{L}_{p} = \mathcal{L}_{CE} (\vz^{\mathcal{S}} / \tau, \vz^{\mathcal{T}} / \tau), 
\end{equation}
where $\vz^{\mathcal{S}}$ and $\vz^{\mathcal{T}}$ are logit vectors produced by student and teacher models, respectively. $\mathcal{L}_{CE}$ is a cross-entropy loss and $\tau$ is a temperature parameter to produce softer probability distributions over classes. We set $\tau$ to $5$ in all our experiments. 
Finally, the final loss objective is defined as: 
\begin{equation}
    \mathcal{L} = 
    \begin{cases}
        \mathcal{L}_a + \mathcal{L}_h & \qquad \triangleright\ \text{Stage I}, \\ 
        \mathcal{L}_p & \qquad \triangleright\ \text{Stage II}. 
    \end{cases}
    \label{eq: total-loss}
\end{equation}
As outlined in the main paper, the total number of epochs is set to $30$, with $15$ epochs allocated to each stage.

\section{Additional Experimental Results} 
\label{appendix: add-experimental-results}

\subsection{Baseline Approaches}

In this section, we describe the baseline methods used for comparison. \textbf{Full fine-tuning} involves updating all parameters of the pre-trained model by initializing them with pre-trained weights and subsequently training them with gradient updates. In federated experiments, this corresponds to the standard FedAvg algorithm \cite{mcmahan2017communication}. \textbf{Linear probing} \cite{alain2017understanding} fine-tunes only the classifier head while keeping the backbone frozen. In federated experiments, only the trainable parameters of the head are shared and averaged. \textbf{LoRA} (Low-Rank Adaptation) \cite{hu2021lora} is a popular parameter-efficient fine-tuning approach, freezing the original weights while training additional low-rank matrices. Its federated counterpart, the FedIT algorithm \cite{zhang2024towards}, applies FedAvg specifically to LoRA parameters. \textbf{Adapter tuning}~\cite{houlsby2019parameter} involves inserting lightweight adapter layers between self-attention (attention) and feed-forward modules, connected via residual connections. Each adapter layer comprises two fully connected layers with biases, separated by a nonlinearity (ReLU activation). 

\begin{figure*}[!bhtp]
    \centering
    \begin{minipage}{0.49\linewidth}
        \centering
        \includegraphics[width=\linewidth]{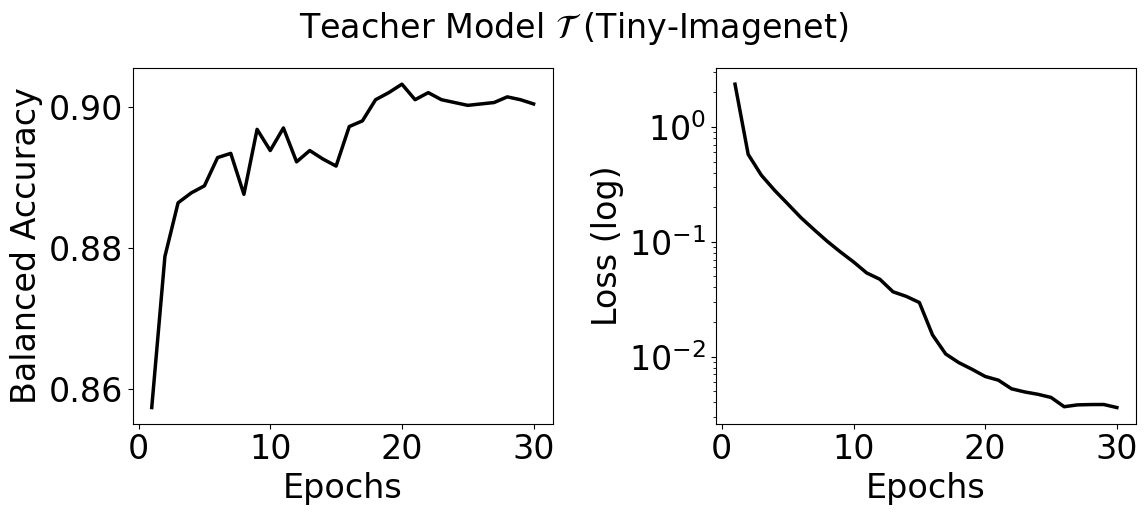}
        \caption{Performance plot of fine-tuning the teacher model on the public auxiliary dataset $\gD_{aux}$ (Tiny-Imagenet). The left plot shows the balanced accuracy, while the right presents the loss performance (log scale).} 
        \label{fig: teacher-imgnet}
    \end{minipage}
    \hfill
    \begin{minipage}{0.49\linewidth}
        \centering
        \includegraphics[width=\linewidth]{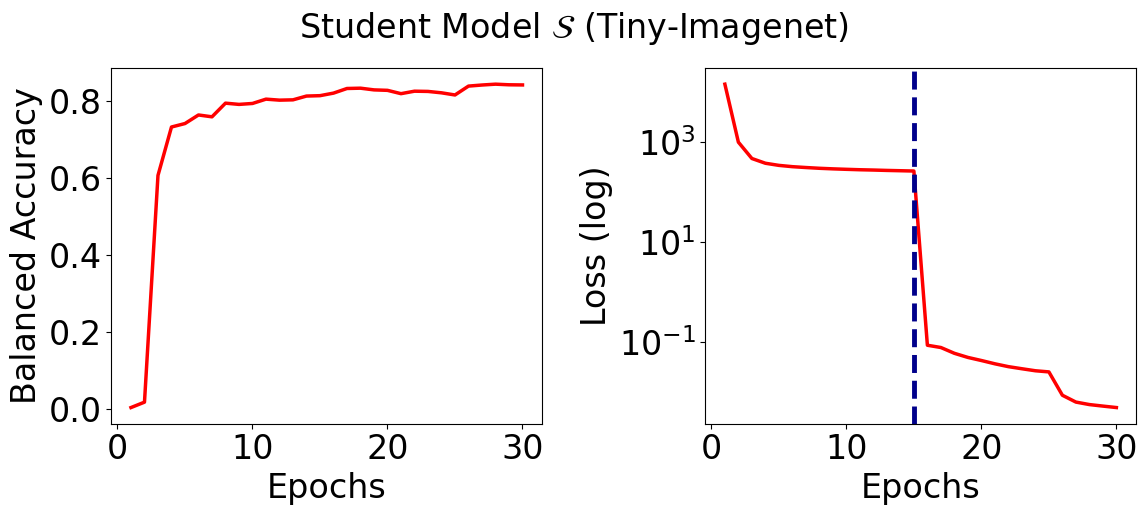}
        \caption{Performance plot of the distillation process on the Tiny-Imagenet dataset. The left plot depicts the balanced accuracy across both stages, while the right shows the loss performance, with the dashed line indicating the beginning of Stage 2 distillation (Eq. \ref{eq: total-loss}).} 
        \label{fig: student-imgnet} 
    \end{minipage}
    
    \vspace{0.5em}

    \begin{minipage}{0.49\linewidth}
        \centering
        \includegraphics[width=\linewidth]{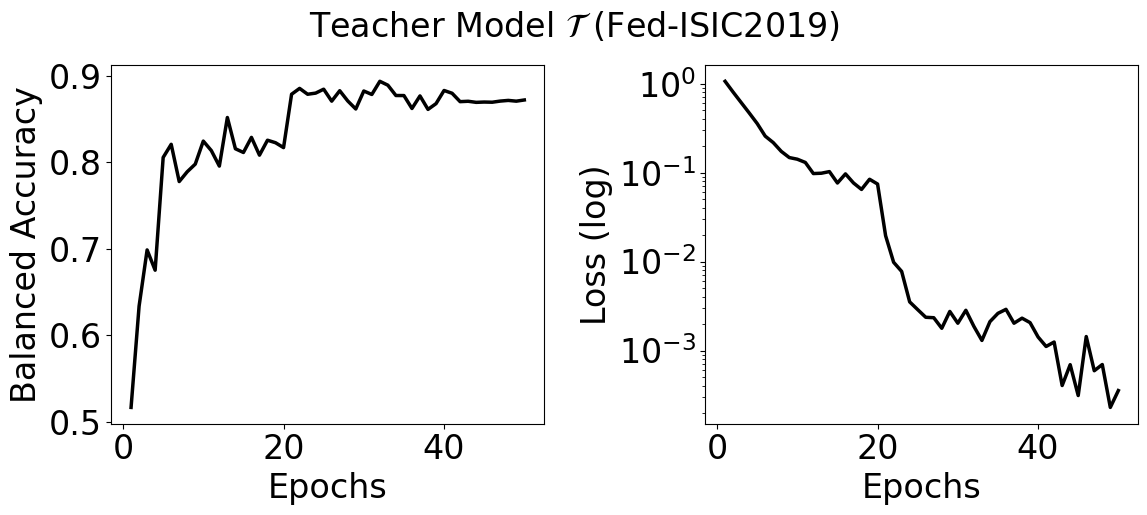}
        \caption{Performance plot of fine-tuning the teacher model on the Fed-ISIC2019 dataset with center=0. The plot on the left shows the balanced accuracy, while the plot on the right presents the loss performance (in a logarithmic scale).}
        \label{fig: teacher-fedisic} 
    \end{minipage}
    \hfill
    \begin{minipage}{0.49\linewidth}
        \centering
        \vspace*{2mm}
        \includegraphics[width=\linewidth]{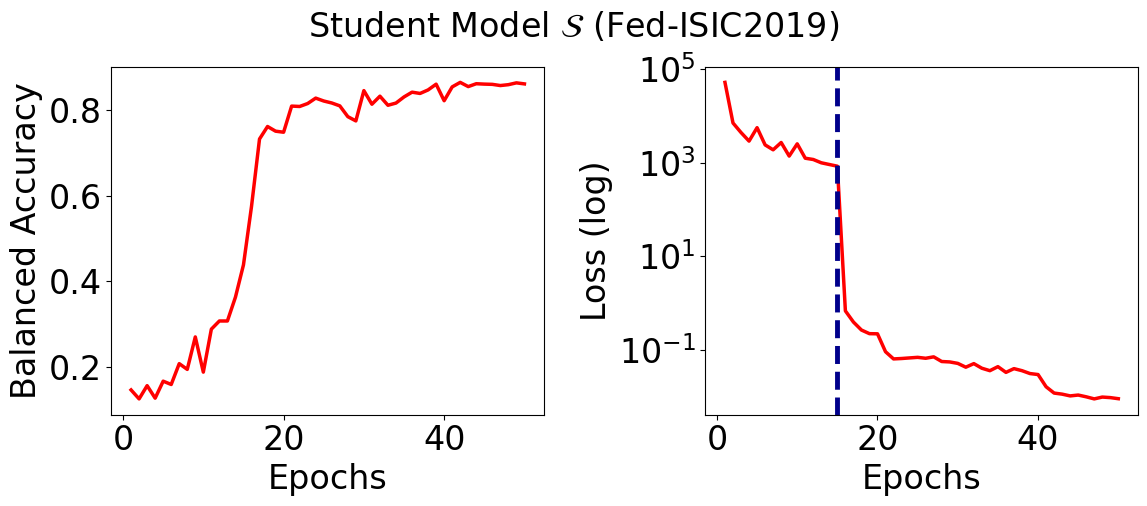}
        \caption{Performance plot of the distillation process on the Fed-ISIC2019 dataset with center=0. The plot on the left depicts the balanced accuracy across both stages, while the right plot shows the loss performance, with the dashed line indicating the beginning of Stage 2 distillation (Eq. \ref{eq: total-loss}).} 
        \label{fig: student-fedisic}
    \end{minipage}
\end{figure*}

\subsection{Main Results} 
\label{appendix: efficiency-results}

Table \ref{tab: main-comparison-params} provides a comprehensive comparison of our method against several baseline approaches, including Full fine-tuning, LoRA, Adapter tuning, and Linear probing, across multiple datasets. This table extends the insights provided in Tables \ref{tab: main-comparison-acc} and \ref{tab: fedisic}, focusing on three key aspects: the number of trainable parameters, latency, and GPU memory requirements.  These results offer a deeper understanding of the trade-offs involved in adapting foundation models for federated learning while maintaining computational feasibility. As observed in the table, full fine-tuning consistently requires a significantly larger number of trainable parameters compared to all other methods, leading to substantially higher memory consumption and latency. In contrast, our method remains efficient, exhibiting a parameter count  close to that of LoRA while requiring even less GPU memory. 

Notably, across all datasets, our approach remains competitive with linear probing in terms of latency while requiring only a marginal increase in parameter storage. Specifically, in the CIFAR-10/SVHN setting, our method achieves near-optimal latency, operating at only a $4\%$ increase over linear probing, despite introducing a significantly greater degree of adaptability. The reduction in trainable parameters compared to LoRA and Adapter tuning further highlights the effectiveness of our approach in maintaining model compactness without sacrificing computational efficiency. 

From a resource perspective, the memory footprint of our method remains consistently low (due to the elimination of the need for backpropagation through the backbone), demonstrating an advantage over more conventional fine-tuning techniques. While full fine-tuning incurs nearly twice the memory cost, our method achieves comparable memory usage to linear probing while outperforming LoRA and Adapter tuning in terms of both efficiency and adaptability. 

These findings underscore the scalability and practicality of our method in federated learning applications. By achieving an optimal balance between parameter efficiency, computational cost, and adaptability, our approach effectively mitigates the overhead typically associated with fine-tuning. Moreover, it demonstrates compatibility with homomorphic encryption techniques, achieving lower computational costs compared to conventional fine-tuning methods.

\begin{table*}
    \centering
    \resizebox{0.875\linewidth}{!}{
        \begin{tabular}{c|ccccc}
        \toprule 
        \textbf{Degree} $d$ & \textbf{Memory (GB)} & \textbf{Latency (s)} & \textbf{Attention loss $(\mathcal{L}_a)$} & \textbf{Representation loss $(\mathcal{L}_h)$} & \textbf{One-epoch accuracy $(\%)$} \\ \midrule 
        $\mathbf{2}$ & $16.82$ & $0.41$ & $44.37$ & $915.46$ & $75.17$ \\ 
        $\mathbf{4}$ & $19.68$ & $0.48$ & $38.31$ & $748.62$ & $83.85$ \\ 
        \rowcolor{lightgray} $\mathbf{6}$ & $22.54$ & $0.52$ & $33.92$ & $655.26$ & $84.36$ \\ 
        $\mathbf{8}$ & $25.39$ & $0.58$ & $30.91$ & $603.53$ & $84.47$ \\ 
        $\mathbf{10}$& $28.99$ & $0.62$ & $28.17$ & $573.54$ & $84.16$ \\ 
        $\mathbf{16}$ & $36.83$ & $0.81$ & $20.31$ & $527.95$ & $85.08$ \\ \midrule 
        $\infty$ \textbf{(True)} & $15.36$ & $0.35$ & $13.59$ & $508.64$ & $86.91$ \\ 
        \bottomrule
        
        \end{tabular}
    }
    \caption{\textbf{Softmax approximation results.} Latency is computed per a batch of samples, when batch size is set to $8$. One-epoch accuracy refers to the accuracy we obtain when performing one full pass over the data for only one epoch. This experiment is carried on the Fed-ISIC2019(center=0) dataset. }
    \label{tab: softmax-approximations}
\end{table*}

\subsection{Distillation Results} 

Following the findings discussed in Section \ref{section: how-distillation}, we present performance plots for two key processes: (i) fine-tuning the teacher model $\mathcal{T}$ on a public auxiliary dataset $\mathcal{D}_{aux}$, and (ii) distilling the student model $\mathcal{S}$ on the same dataset $\mathcal{D}_{aux}$ using the trained teacher model $\mathcal{T}$. Figure \ref{fig: teacher-imgnet} illustrates the performance metrics $-$ balanced accuracy and loss $-$ of the fine-tuning process on the Tiny-ImageNet dataset. After completing this training, we proceed with distillation for the approximation-enabled student model. Figure \ref{fig: student-imgnet} shows the performance plot for the distillation process on the Tiny-ImageNet dataset. In the loss behavior plot (on the right), a dashed line marks the start of the second stage. During the first stage, as described in Eq. \ref{eq: total-loss}, transformer-layer distillation is performed (showing a higher value due to the high-dimensional attention and hidden representation matrices). Then, the second-stage distillation follows, which is a prediction-layer distillation \cite{hinton2015distilling}. The learning rate scheduler is employed at epochs $15$ and $25$ with a decay factor of $0.1$. Figures \ref{fig: teacher-fedisic} and \ref{fig: student-fedisic} present analogous performance plots for the Fed-ISIC2019 dataset, using center=0 as the public auxiliary dataset $\mathcal{D}_{aux}$. The learning rate scheduler is employed at epochs $20$ and $40$ with a decay factor of $0.1$.

\subsection{Approximation Results} 
In this section, we revisit and elaborate on the approximations introduced in the main paper. 

\vspace{0.75em} 
\noindent\textbf{Softmax.} Given a vector $\vz \in \mathbb{R}^n$ with components $z_i$, the softmax function is defined as: 
\begin{equation*}
    \text{softmax}(z_i) = \dfrac{e^{z_i}}{\sum_{j=1}^n e^{z_j}}
\end{equation*}
where $e^{z_i}$ denotes the exponential function applied to the component $z_i$. The goal is to approximate the softmax function using a polynomial approximation of the exponential function, $P_n(\vz_i)$, and to estimate the maximum deviation in softmax values due to this approximation.

\noindent The Taylor series provides a polynomial approximation of $e^x$ around $x=0$ up to $d$ terms: 
\begin{equation*}
    e^x \approx P_d(x) = \sum_{k=0}^d \dfrac{x^k}{k!} 
\end{equation*}
The error bound of this approximation is given by the remainder term $R_n(x)$, expressed as: 
\begin{equation*}
    R_d(x) = \dfrac{e^{\xi}}{(d+1)!} x^{d+1} 
\end{equation*}
for some $\xi$ between $0$ and $x$, indicating the error in approximating $e^x$ with $P_d(x)$.

\begin{table*}[t]
    \centering
    \resizebox{0.875\linewidth}{!}{
        \begin{tabular}{c|ccccc}
            \toprule 
             \textbf{Degree} $d$ & \textbf{Memory (GB)} & \textbf{Latency (ms)} & \textbf{Attention loss $(\mathcal{L}_a)$} & \textbf{Representation loss $(\mathcal{L}_h)$} & \textbf{One-epoch accuracy $(\%)$} \\ \midrule 
                                   \textbf{2} & $22.42$ & $118.70$ & $61.50$ & $1698.19$ & $57.87$ \\ 
                                   \textbf{4} & $22.44$ & $119.64$ & $57.42$ & $1534.97$ & $76.64$ \\ 
                                   \textbf{6} & $22.45$ & $120.27$ & $45.67$ & $1352.91$ & $81.77$ \\ 
              \rowcolor{lightgray} \textbf{7} & $22.46$ & $120.43$ & $44.11$ & $1334.32$ & $82.96$ \\ 
                                   \textbf{8} & $22.47$ & $120.97$ & $45.10$ & $1352.55$ & $83.23$ \\ 
                                  \textbf{10} & $22.48$ & $121.70$ & $43.23$ & $1332.62$ & $80.61$ \\ 
                                  \textbf{12} & $22.50$ & $122.98$ & $40.40$ & $1302.41$ & $82.55$ \\ 
                                  \textbf{16} & $22.53$ & $124.76$ & $38.86$ & $1292.36$ & $83.04$ \\ \midrule
                     $\infty$ \textbf{(True)} & $22.46$ & $\,\,\,10.57$ & $38.77$ & $1289.71$ & $83.15$ \\ 
             \bottomrule
        \end{tabular}
    }
    \caption{\textbf{Approximation of the reciprocal (inverse function).} Latency is computed per a batch of samples (aggregate over $L$ transformer blocks), when batch size is set to $8$. One-epoch accuracy refers to the accuracy we obtain when performing one full pass over the data for only one epoch. This experiment is carried on the Fed-ISIC2019 (center=0) dataset.} 
    \label{tab: inverse-approximations}
\end{table*}

\vspace{0.5em}
\noindent\textbf{Softmax approximation results.} Table \ref{tab: softmax-approximations} presents softmax approximations while keeping all other non-linear components fixed. Using Eq. \ref{eq: exp-approx}, we vary the polynomial degree $d$ for the approximation during a single epoch of knowledge distillation (Section \ref{section: how-distillation}, Stage I) and compare the results to the true softmax. The table reports key performance metrics, including attention loss $\mathcal{L}_a$ (Eq. \ref{eq: attn}), representation loss $\mathcal{L}_h$ (Eq. \ref{eq: hidd2}), and accuracy. Additionally, it provides efficiency metrics such as GPU memory usage and latency for processing a batch of samples. Balancing the trade-offs between time, memory requirements, and performance drop, we chose $d=6$ for all experiments. 

\vspace{0.5em}
\noindent\textbf{Inverse.} Since homomorphic encryption supports only addition and multiplication operations, it cannot perform division directly. To make it work in the encryption domain, we approximate the inverse function. The details of this approximation are presented in Eq. \ref{eq: division}, and the corresponding algorithm is outlined in Algorithm \ref{algorithm: inverse}. To ensure that  the value of $x$ falls within the range $(0,2)$, we determine the maximum possible value of $x$ in the plaintext domain and scale it accordingly in the encrypted domain using this maximum value. 

\begin{algorithm}[t]
    \caption{Inverse algorithm}\label{algorithm: inverse} 
    \textbf{Input:} $0 < x < 2, \text{ degree } d \in \mathbb{N}$ \\ 
    \textbf{Output:} an approximation of $1/x$ (Eq. \ref{eq: division})
    \begin{algorithmic}[1]
        \State $a_0 \gets 2 - x$ 
        \State $b_0 \gets 1 - x$ 
        \For{$n \gets 0$ to $d-1$}
            \State $b_{n+1} \gets b_n^2$ 
            \State $a_{n+1} \gets a_n \cdot (1 + b_{n+1})$
        \EndFor 
        \State \Return $a_d$ 
    \end{algorithmic}
\end{algorithm}

\vspace{0.5em}
\noindent \textbf{Inverse approximation results.} Table \ref{tab: inverse-approximations} summarizes the results of inverse function approximations, keeping all other non-linear components fixed and utilizing the softmax approximation with a $6$th-degree polynomial. Following Algorithm \ref{algorithm: inverse} and Eq. \ref{eq: division}, we varied the polynomial degree $d$ under the same conditions described above. While the memory and time requirements remain nearly identical across different values of $d$, the focus is on performance metrics. Based on these considerations, we selected $d=7$, as it closely approximates the true inverse in terms of losses and accuracy.

\begin{table}[t]
    \centering
    \resizebox{\linewidth}{!}{
    \begin{tabular}{ll}
        \toprule
         \textbf{Time taken to encrypt one sample} & $1062$ ms \\ \midrule 
         \textbf{Time taken to decrypt one sample} & $168.7$ ms \\ \midrule 
         \textbf{Time taken to encrypt a batch of 8 samples} & $4.65$ s \\ \midrule 
         \textbf{Time taken to decrypt a batch of 8 samples} & $231.59$ ms \\ \midrule 
         \textbf{Ciphertext size of one sample} & $17.33$ MB \\ \midrule
         \textbf{Plaintext size of one sample} & $6.21$ MB \\ \midrule
         \textbf{Encrypted inference time} & $136$ s \\ 
         \bottomrule 
    \end{tabular}
    }
    \caption{Computational and memory overhead for encrypted inference and encryption using FHE with the Tile Tensors framework. Results are presented per transformer block, evaluated at a 128-bit security level, including approximations for Softmax, inverse, and GELU operations. }
    \label{tab: fhe-implementation-results}
    \vskip -1em 
\end{table}

\subsection{Homomorphic Encryption Results}
\label{appendix: homomorphic}
For the implementation of fully homomorphic encryption (FHE), we utilize the Tile Tensors framework \cite{aharoni2023helayers, cheon2017homomorphic} on an NVIDIA A100-PCIE-40GB machine with 250 GB of available RAM. Table \ref{tab: fhe-implementation-results} summarizes the computational and memory requirements for our experiments, all evaluated at the standard 128-bit security level. The results presented in the table correspond to the encryption of one transformer block, accounting for approximations of operations such as Softmax, inverse, and GELU. Specifically, we report both the encrypted inference time and ciphertext memory size. The encrypted inference is executed on a powerful server, and the required time determines its computational complexity and latency, and the ciphertext size directly determines the communication overhead within the federated setting. 

The encryption of a single data sample with a dimension $(577, 768)$ takes approximately $1062$ ms, while decrypting the same sample takes significantly less time, around $168.7$ ms, highlighting the asymmetry in encryption and decryption costs. Batch processing improves efficiency, as encrypting $8$ samples takes $4.65$ seconds, whereas decryption for the same sized batch completes in $231.59$ ms. The ciphertext size of a single encrypted sample expands to $17.33$ MB, larger than its plaintext counterpart ($6.21$ MB), indicating the additional communication overhead introduced by encryption. Moreover, encrypted inference for one transformer block requires approximately $136$ seconds on the specified hardware, highlighting the computational overhead of inference under encryption. Nonetheless, this inference time could be substantially reduced with optimized cryptographic primitives, more powerful server infrastructure, and GPU acceleration \cite{kim2024cheddar, jin2024efficient, yang2024phantom, papadakis2024towards}. 

\vspace{1em} 
\noindent\textbf{Computational Overhead.} As with any cryptographic approach for privacy-preserving ML, BlindFed introduces non-negligible computational costs. However, it is designed to minimize the burden on thin clients by offloading most computations to a powerful server (Table~\ref{tab: fhe-implementation-results}). On the client side, only encryption/decryption of intermediate representations and a plaintext parallel adapter update are performed. On commodity CPUs, encrypting (decrypting) a single sample takes $\mathbf{1.06\,s}$ ($\mathbf{0.17\,s}$), and encrypting (decrypting) $8$ samples in batch takes $\mathbf{4.7\,s}$ ($\mathbf{0.23\,s}$), requiring less than $\mathbf{1\,GB}$ of RAM. The server handles the expensive encrypted inference for each ViT block under CKKS, taking roughly $\mathbf{136\,s}$ per sample (amortized) and requiring more than $\mathbf{22\,GB}$ of memory. However, as mentioned above, these operations are embarrassingly parallel across clients, and GPU-accelerated CKKS implementations can further reduce the server-side inference time by over $10\times$.

\end{document}